\documentclass{article}
\pdfoutput=1
\usepackage[preprint,nonatbib]{neurips_2020}

\usepackage[utf8]{inputenc} % allow utf-8 input
\usepackage[T1]{fontenc}    % use 8-bit T1 fonts
\usepackage{hyperref}       % hyperlinks
\usepackage{url}            % simple URL typesetting
\usepackage{booktabs}       % professional-quality tables
\usepackage{amsfonts}       % blackboard math symbols
\usepackage{nicefrac}       % compact symbols for 1/2, etc.
\usepackage{microtype}      % microtypography

\title{Region-based Energy Neural Network for Approximate Inference}

\author{%
  Dong Liu, Ragnar Thobaben, Lars K. Rasmussen% \thanks{Use footnote for providing further information
  \\
  Division of Information Science and Engineering\\
  KTH Royal Institute of Technology\\
  Stockholm, Sweden \\
  \texttt{\{doli, ragnart, lkra\}@kth.se} \\
}

\usepackage{mystyle}
\usepackage{xr}
\usepackage{subcaption}
\usepackage{tikz}

\usetikzlibrary{calc,shapes,positioning,fit}
\usetikzlibrary{arrows}

\pdfoutput=1
\usepackage{adjustbox}
\begin{document}

\maketitle
\begin{abstract}
Region-based free energy was originally proposed for generalized belief propagation (GBP) to improve loopy belief propagation (loopy BP). In this paper, we propose a neural network based energy model for inference in general Markov random fields (MRFs), which directly minimizes the region-based free energy defined on region graphs. We term our model  Region-based Energy Neural Network (RENN). Unlike message-passing algorithms, RENN avoids iterative message propagation and is faster. Also different from recent deep neural network based models, inference by RENN does not require sampling, and RENN works on general MRFs. RENN can also be employed for MRF learning. Our experiments on marginal distribution estimation, partition function estimation, and learning of MRFs show that RENN outperforms the mean field method, loopy BP, GBP, and the state-of-the-art neural network based model.
\end{abstract}

\section{Introduction}
Probabilistic graphical models offer a natural way of encoding conditional dependencies of random variables. Message-passing algorithms are practical and powerful methods to solve probabilistic inference problems on graphical models, including inferring the overall state of a system or marginal distributions
of subsets of nodes in the system. The well-known standard belief propagation (BP) algorithm \cite{Pearl1982reverend,kschischang2001factor_graph} has been popularly used in exact inference problems on tree-structured graphs and approximates inference in general graphs (i.e., loopy BP), which was explained by the Bethe free energy minimization later on \cite{yedidia2003understanding}. The approximate inference of BP was then improved by the generalized BP (GBP, also known as the parent-to-child algorithm), which is also an iterative message-passing algorithm on a constructed region graph \cite{Yedidia:2000:GBP:3008751.3008848, yedida2005constucting}. GBP propagates messages between regions (i.e., clusters of nodes) and is generally more accurate than loopy BP. Fixed points of GBP that operates on region graphs, correspond to stationary points of the region-based free energy of the region graphs. Depending on the graph size and potential functions, the iterative message-passing algorithms can take a long time to converge before returning inference results (if they can converge at all). Also, inference of these message-passing methods can degenerate significantly in dense graphs.

Recent deep generative models \cite{DBLP:journals/corr/KingmaW13,2017arXiv170104722M, 2017arXiv171101558T,li2018graphical, johansonNIPS2016_6379} show promising results on directed graphical models and for pre-defined inference tasks such as a posteriori estimation of latent variables.
These models are advantageously fast on modern GPUs, but mainly perform directed graphical modeling and usually do not explicitly and fully model the dependencies of structured random variables. End-to-end training is used to learn a generative network, and they also require a separate neural network 
for recognition (i.e., pre-defined inference task). Sampling is usually required to perform neural network training and the pre-defined inference.

In this paper, we proposed a model to combine the benefits of both and to avoid the drawbacks of each. Specifically, we use a neural network to directly minimize the region-based free energy for general approximate probabilistic inference in MRFs (instead of a pre-defined inference task) and without iterative message-passing as belief propagation methods do. We term the region-based energy neural network RENN. RENN allows quick approximate inference and outperforms loopy BP, GBP, and the state-of-the-art neural network based inference model. The advantages of RENN remain even in challenging complete graphs where every two nodes are connected. We also consider learning MRFs by using RENN for inference. Learning with RENN outperforms benchmark methods.
In neither learning MRFs with RENN nor employing RENN for inference only, sampling is required.

\begin{figure}[!t]
  \centering
  \begin{tikzpicture}
    \begin{scope}[scale=0.7]

    \tikzstyle{cnode} = [thick, draw=black, circle, inner sep = 1pt,  align=center]
    \tikzstyle{nnode} = [thick, rectangle, rounded corners = 0pt,draw,inner sep = 2pt]
    \node[cnode] (x1) at (0,0) {1};
    \node[cnode] (x2) at (2,0) {2};
    \node[cnode] (x3) at (4,0) {3};

    \node[cnode] (x4) at (0,-2) {4};
    \node[cnode] (x5) at (2,-2) {5};
    \node[cnode] (x6) at (4,-2) {6};

    \node[nnode] (fa) at (1,0) {\small$A$};
    \node[nnode] (fb) at (3,0) {\small$B$};

    \node[nnode] (fc) at (0,-1) {\small$C$};
    \node[nnode] (fd) at (2,-1) {\small$D$};
    \node[nnode] (fe) at (4,-1) {\small$E$};
        
    \node[nnode] (ff) at (1,-2) {\small$F$};
    \node[nnode] (fg) at (3,-2) {\small$G$};

    \draw[-] (x1) -- (fa);
    \draw[-] (x1) -- (fc);

    \draw[-] (x2) -- (fa);
    \draw[-] (x2) -- (fb);
    \draw[-] (x2) -- (fd);

    \draw[-] (x3) -- (fb);
    \draw[-] (x3) -- (fe);

    \draw[-] (x4) -- (fc);
    \draw[-] (x4) -- (ff);

    \draw[-] (x5) -- (fd);
    \draw[-] (x5) -- (ff);
    \draw[-] (x5) -- (fg);

    \draw[-] (x6) -- (fe);
    \draw[-] (x6) -- (fg);

    \end{scope}
    \draw[dashed] (0, -1.8) -- (4.0, -1.8) (4, 0) -- (4, -3.5);
    
    \begin{scope}[xshift=0.5cm, yshift=-2.35cm,scale=0.7]
      \tikzstyle{rnode} = [thick, rectangle, rounded corners = 2pt,minimum size = 0.0cm,draw,inner sep = 2pt]
      \node[rnode] (r01) at (0,0) {\small \begin{tabular}[x]{@{}c@{}}1, 2, 4, 5 \\ $A,C,D,F$ \end{tabular}};
      \node[rnode] (r02) at (3,0) {\small \begin{tabular}[x]{@{}c@{}}2, 3, 5, 6\\ $B,D,E,G$ \end{tabular}};
      \node[rnode] (r11) at (1.5, -1.5) {\small 2, 5, $D$};

      \draw[->] (r01.south) -- (r11.north);
      \draw[->] (r02.south) -- (r11.north);

    \end{scope}

    \begin{scope}[xshift=5.3cm, yshift=-0.50cm,scale=0.7]
      \tikzstyle{rnode} = [thick, rectangle, rounded corners = 2pt,minimum size = 0.0cm,draw,inner sep = 2pt]
      \node[rnode] (r01) at (0,0) {\small\begin{tabular}[x]{@{}c@{}}1, 2, 4, 5 \\ $A,C,D,F$ \end{tabular}};
      \node[rnode] (r02) at (3.7,0) {\small\begin{tabular}[x]{@{}c@{}}2, 3, 5, 6\\ $B,D,E,G$ \end{tabular}};
      \node[rnode] (r03) at (8,0) {\small\begin{tabular}[x]{@{}c@{}}1, 2, 3, 4, 5, 6\\ $A,B,C,E,F,G$ \end{tabular}};
      \begin{scope}[yshift=-0.2cm]
      \node[rnode] (r11) at (0, -2.0) {\small\begin{tabular}[x]{@{}c@{}}1, 2\\ $A$ \end{tabular}};
      \node[rnode] (r12) at (1.5, -2.0) {\small\begin{tabular}[x]{@{}c@{}}2, 3\\ $B$ \end{tabular}};
      \node[rnode] (r13) at (3, -2.0) {\small\begin{tabular}[x]{@{}c@{}}1, 4\\ $C$ \end{tabular}};
      \node[rnode] (r14) at (4.5, -2.0) {\small\begin{tabular}[x]{@{}c@{}}2, 5\\ $D$ \end{tabular}};
      \node[rnode] (r15) at (6, -2.0) {\small\begin{tabular}[x]{@{}c@{}}3, 6\\ $E$ \end{tabular}};
      \node[rnode] (r16) at (7.5, -2.0) {\small\begin{tabular}[x]{@{}c@{}}4, 5\\ $F$ \end{tabular}};
      \node[rnode] (r17) at (9, -2.0) {\small\begin{tabular}[x]{@{}c@{}}5, 6\\ $G$ \end{tabular}};

      \begin{scope}[yshift=0.5cm]
      \node[rnode] (r21) at (1, -4) {\small 1};
      \node[rnode] (r22) at (2.5, -4) {\small 2};
      \node[rnode] (r23) at (4, -4) {\small 3};
      \node[rnode] (r24) at (5.5, -4) {\small 4};
      \node[rnode] (r25) at (7, -4) {\small 5};
      \node[rnode] (r26) at (8.5, -4) {\small 6};
      \end{scope}
      \end{scope}
      % edge level0 to level1
      \draw[->] (r01.south) -- (r11.north);
      \draw[->] (r03.south) -- (r11.north);

      \draw[->] (r02.south) -- (r12.north);
      \draw[->] (r03.south) -- (r12.north);

      \draw[->] (r01.south) -- (r13.north);
      \draw[->] (r03.south) -- (r13.north);

      \draw[->] (r01.south) -- (r14.north);
      \draw[->] (r02.south) -- (r14.north);

      \draw[->] (r02.south) -- (r15.north);
      \draw[->] (r03.south) -- (r15.north);

      \draw[->] (r01.south) -- (r16.north);
      \draw[->] (r03.south) -- (r16.north);

      \draw[->] (r02.south) -- (r17.north);
      \draw[->] (r03.south) -- (r17.north);

      % edge level1 to level2
      \draw[->] (r11.south) -- (r21.north);
      \draw[->] (r13.south) -- (r21.north);

      \draw[->] (r11.south) -- (r22.north);
      \draw[->] (r12.south) -- (r22.north);
      \draw[->] (r14.south) -- (r22.north);

      \draw[->] (r12.south) -- (r23.north);
      \draw[->] (r15.south) -- (r23.north);

      \draw[->] (r13.south) -- (r24.north);
      \draw[->] (r16.south) -- (r24.north);

      \draw[->] (r14.south) -- (r25.north);
      \draw[->] (r16.south) -- (r25.north);
      \draw[->] (r17.south) -- (r25.north);

      \draw[->] (r15.south) -- (r26.north);
      \draw[->] (r17.south) -- (r26.north);
    \end{scope}
  \end{tikzpicture}
  % \vskip -0.15in
  \caption{Illustration of a factor graph for 2-by-3 grid (top left, variable nodes are indexed by number and factor nodes by letters), and two alternative regions graphs (two levels for the bottom-left one and three levels for right one) constructed from the factor graph.}
  \label{fig:factor-region-graphs}
  \vskip -0.2in
\end{figure}
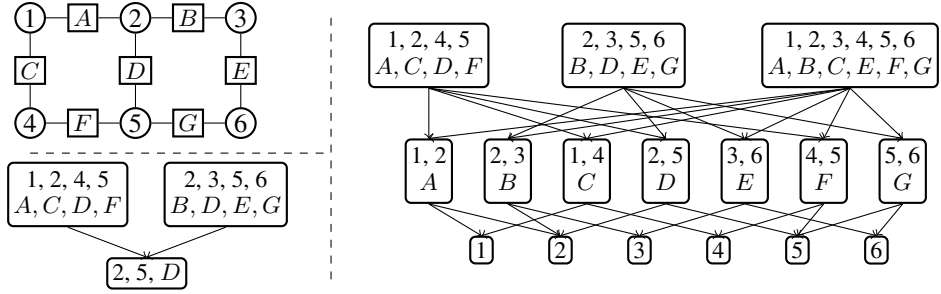
\section{Preliminaries}
\label{sec:preliminary}
Let $\left\{X_1,\cdots, X_N\right\}$ be a set of $N$ discrete-valued random variables and $x_i$ represent the possible realization of $X_i$. We denote the joint probability function $p(X_1=x_1, \cdots, X_N=x_N; \bm{\theta})$ parameterized by $\bm{\theta}$, by $p(\bm{x}; \bm{\theta})$. To keep the notation simple, we define that each variable has $K$ states.

In a MRF, the joint probability distribution of $\bm{x}$ can be written as
\begin{equation}\label{eq:joint-px}
  p(\bm{x}; \bm{\theta}) = \frac{1}{Z(\bm{\theta})} \prod_{a\in \Ff} \psi_a(\bm{x}_a; \bm{\theta}_a),
\end{equation}
where $a$ indexes potential functions in set $\Ff$. The potential function $\psi_a(\bm{x}_a;\bm{\theta}_a)$ is nonnegative, parameterized by $\bm{\theta}_a$, and has arguments $\bm{x}_a$ that are some subset of $\left\{ x_1, x_2, \cdots, x_N \right\}$. $Z(\bm{\theta})$ is the partition function of $p(\bm{x}; \bm{\theta})$, i.e. $Z(\bm{\theta}) = \sum_{\bm{x}}\prod_{a} \psi_a(\bm{x}_a;\bm{\theta}_a)$.

\begin{defn}[Factor Graph]\label{def:factor-graph}
A factor graph $\Gg_F$, is a bipartite graph that represents the factorization structure of \eqref{eq:joint-px}. A factor graph has two types of nodes: i) a variable node for each variable $x_i$; ii) a factor node for each potential function $\psi_a$. An edge exists between a variable node $i$ and factor node $a$ if and only if $x_i$ is in the argument of $\psi_a$. We denote a factor graph by $\Gg_F(\Vv \cup \Ff, \Ee_F)$ with  the set of variable nodes $\Vv$,  the set of factor nodes $\Ff$, and  the set of undirected edges $\Ee_F$.
\end{defn}

Loopy BP as a message-passing algorithm operates on factor graphs (see, e.g., the top-left example in Figure~\ref{fig:factor-region-graphs}) and computes the marginal distributions of \eqref{eq:joint-px}. These estimations are done by iteratively exchanging messages between factor and variable nodes in a factor graph ({see Appendix~\ref{apdix:sec:bethe-lBP} for detailed discussions and message update rules}). Loopy BP has the interpretation of minimizing the well-known Bethe free energy \cite{yedidia2003understanding}
\begin{align}\label{eq:bethe-free-energy}
  F_{B} = \sum_{a\in \Ff} \sum_{\bm{x}_a} b_a(\bm{x}_a)\ln{\frac{b_a(\bm{x}_a)}{\psi_a(\bm{x}_a)}} -  \sum_{i=1}^{N} (d_i - 1) \sum_{x_i} b_i(x_i) \ln{b_i(x_i)},
\end{align}
where $d_i$ is the degree of node $i$ in the underlying factor graph  (i.e., the number of neighbors of node $i$), $b_a(\bm{x}_a)$ and $b_i({x}_i)$ are beliefs (marginal probability estimations) for $\bm{x}_a$ and $x_i$, respectively. The free energy interpretation connects the Kullback-Leibler (KL) divergence to the (loopy) BP algorithm (see Appendix~\ref{apdx:sec:variational-free-energy-and-mf} and \ref{apdix:sec:bethe-lBP}). Importantly, for a tree-structured underline graph of $p(\bm{x}; \bm{\theta})$, $\min_{\left\{ b_a, a\in \Ff \right\}}{F_{B}} = - \log{Z(\bm{\theta})}$ \cite{koller2009pgm, Bishop:2006:PRM:1162264}, in which case Bethe free energy is precisely $\mathrm{KL}(b(\bm{x})\|p(\bm{x};\bm{\theta})) - \log{Z(\bm{\theta})}$ with $b(\bm{x})$ as a variational distribution that marginalizes to $\left\{ b_a ,b_i\right\}$. For general graphs containing loops, $\min_{\left\{ b_a, a\in \Ff \right\}}{F_{B}}$ gives an approximation to $-\log{Z(\bm{\theta})}$ \cite{wainwright2008graphical, weller2014approximating, weller2014understanding}.

Apart from loopy BP \cite{yedidia2003understanding} and its variants \cite{roosta2008reweighed_sum_product, Pretti2005damping, liu2019alpha} that are iterative message-passing algorithms, an alternative way for general inference for marginalization and partition in MRFs is to directly solve the Bethe free energy minimization problem by a gradient descent method \cite{welling2001belief,xionggyr19one-shot,NIPS2019_9687}. For instance, \cite{welling2001belief} updates the marginals of univariate variables by a gradient descent method. \cite{NIPS2019_9687} generalizes this approach by updating all marginals in an MRF by minimizing the Bethe free energy, where the marginals are amortized by a neural network, leading to the Inference Net model.

The Bethe approximation is restricted to the factorization form of $p(\bm{x};\bm{\theta})$ and has a poor estimation performance  in dense graphs or loops with conflicting potential preferences \cite{koller2009pgm}. GBP was developed to overcome this limitation and to improve the performance of loopy BP, in which the messages are propagated among sets of nodes or regions (see Appendix~\ref{apdix:sec:gbp}). A region graph is a structured graph that originally was proposed to organize the computation of GBP messages. Two alternative region graphs constructed from the same factor graph are shown in Figure~\ref{fig:factor-region-graphs}. Region graphs give us  the freedom to customize   how we cluster  nodes in a factor graph into a region graph. Apart from the flexibility, if a loop with conflicting potentials in a factor graph is cast into a region, the above-mentioned difficulty can be circumvented naturally. A region graph is formally defined as:
\begin{defn}[Region Graph]\label{def:region-graph}
  A \textit{region} $R$ consists of a set $V_R$ of variables nodes and a set $A_R$ of factor nodes such that if a factor node $a$ belongs to $A_R$, all the variables nodes neighboring $a$ are in $V_R$.
  A \textit{region graph} is  a directed graph $\Gg_R(\Rr, \Ee)$, where each vertex $R \in \Rr$ is defined as the joint set of variable and factor nodes in this region, i.e. $R = \left\{ i \in V_R, a \in A_R | i \in \Vv, a \in \Ff \right\}$. Each edge $e \in \Ee$ in $\Gg_R$ is directed from $R_p$ to $R_c$ such that $R_c \subset R_p$. 
\end{defn}
We can associate the \textit{Region-based free energy} with a region graph, which plays a similar role as the Bethe free energy for a factor graph.
\begin{defn}[Region-based Free Energy]\label{def:region-free-energy}
  Given a region $R$ in $\Gg$ and $\bm{\theta}_R=\{\bm{\theta}_a, a\in A_R\}$, the region energy is defined to be $E_R(\bm{x}_R; \bm{\theta}_R) = - \sum_{a\in A_R} \ln{\psi_a(\bm{x}_a; \bm{\theta}_a)}$. For any region graph $\Gg_R$, the region-based free energy is defined as
  \begin{equation}\label{eq:def-region-free-energy}
    F_R(\Bb; \bm{\theta}) = \hspace{-0.15cm}\sum_{R\in \Rr} \hspace{-0.1cm}c_R\hspace{-0.1cm} \sum_{\bm{x}_R}b_R(\bm{x}_R) (E_R(\bm{x}_R; \bm{\theta}_R) + \ln{b_R}(\bm{x}_R)),
  \end{equation}
  where $b_R(\bm{x}_R)$ is the belief to region $R$, $\Bb$ is the set of region beliefs $\Bb = \left\{ b_R| R \in \Rr \right\}$, and the integer $c_R \in \NN$ is the counting number for region $R$.
\end{defn}
{The minimized region-based free energy equals to the negative log-partition function of $p(\bm{x};\bm{\theta})$, i.e. $\min_{\Bb}F_R(\Bb;\bm{\theta}) = -\log{Z(\bm{\theta})}$, if each belief is exactly the corresponding marginalization, $b_R(\bm{x}_R)=p(\bm{x}_{R})$, $\forall~R\in \Rr$ \cite{yedidia2003understanding,yedida2005constucting}. Otherwise, $\min_{\Bb}F_R(\Bb;\bm{\theta})$ is an approximation of $\min_{\Bb}F_R(\Bb;\bm{\theta}) \approx -\log{Z(\bm{\theta})}$ for general cases.}

\section{Region-based Energy Neural Network}
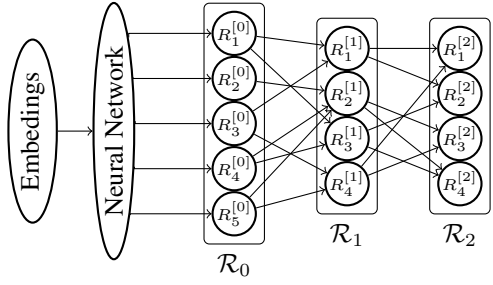
\begin{figure}[!t]
  \centering
\begin{tikzpicture}
    \tikzstyle{enode} = [thick, draw=black, ellipse, inner sep = 2pt,  align=center]
    \tikzstyle{cnode} = [thick, draw=black, circle, inner sep = 0.0pt,  align=center]
    \tikzstyle{nnode} = [thick, rectangle, rounded corners = 0pt,draw,inner sep = 2pt]
    \begin{scope}[xshift=-4cm, yshift=0cm, scale=1.0]

    \begin{scope}[xshift=-1cm, yshift=-1.3cm, scale=0.6]
      \node[enode, rotate=90] (em) at (-1.8,0) {Embedings};
      \node[enode, rotate=90] (nn) {Neural Network};
    \end{scope}
    % level0 regions
    \begin{scope}[scale=0.6]
    \node[cnode] (r01) at (1, 0) {\tiny$R_1^{[0]}$};
    \node[cnode] (r02) at (1, -1) {\tiny$R_2^{[0]}$};
    \node[cnode] (r03) at (1, -2) {\tiny$R_3^{[0]}$};
    \node[cnode] (r04) at (1, -3) {\tiny$R_4^{[0]}$};
    \node[cnode] (r05) at (1, -4) {\tiny$R_5^{[0]}$};
    \node[label=below:$\Rr_0$, draw,rounded corners = 2pt, inner sep=1mm, fit=(r01) (r05)] {};
  \end{scope}

  \draw[->] (nn.351.9) |- (r01);
  \draw[->] (nn.340) |- (r02);
  \draw[->] (nn.295) |- (r03);
  \draw[->] (nn.210) |- (r04);
  \draw[->] (nn.191) |- (r05);
  \draw[->] (em) -- (nn);

    % level 1 regions
    \begin{scope}[xshift=1.5cm, yshift=-0.2cm, scale=0.6]
    \node[cnode] (r11) at (1, 0) {\tiny$R_1^{[1]}$};
    \node[cnode] (r12) at (1, -1) {\tiny$R_2^{[1]}$};
    \node[cnode] (r13) at (1, -2) {\tiny$R_3^{[1]}$};
    \node[cnode] (r14) at (1, -3) {\tiny$R_4^{[1]}$};
    \node[label=below:$\Rr_1$, draw, rounded corners = 2pt, inner sep=1mm, fit=(r11) (r14)] {};
    \end{scope}

    % level 1 regions
    \begin{scope}[xshift=3cm, yshift=-0.2cm, scale=0.6]
    \node[cnode] (r21) at (1, 0) {\tiny$R_1^{[2]}$};
    \node[cnode] (r22) at (1, -1) {\tiny$R_2^{[2]}$};
    \node[cnode] (r23) at (1, -2) {\tiny$R_3^{[2]}$};
    \node[cnode] (r24) at (1, -3) {\tiny$R_4^{[2]}$};
    \node[label=below:$\Rr_2$, draw, rounded corners = 2pt, inner sep=1mm, fit=(r21) (r24)] {};
    \end{scope}

    \draw[->] (r01) -- (r11);
    \draw[->] (r03) -- (r11);

    \draw[->] (r02) -- (r12);
    \draw[->] (r04) -- (r12);
    \draw[->] (r05) -- (r12);

    \draw[->] (r01) -- (r13);
    \draw[->] (r04) -- (r13);

    \draw[->] (r03) -- (r14);
    \draw[->] (r05) -- (r14);

    \draw[->] (r11) -- (r21);
    \draw[->] (r14) -- (r21);

    \draw[->] (r11) -- (r22);
    \draw[->] (r13) -- (r22);

    \draw[->] (r12) -- (r23);
    \draw[->] (r14) -- (r23);

    \draw[->] (r12) -- (r24);
    \draw[->] (r13) -- (r24);

    \end{scope}
  \end{tikzpicture}
  \caption{Illustration of a RENN with three levels of regions ($\Rr_0$, $\Rr_1$, $\Rr_2$).}  \label{fig:renn-illustration}
  \vskip -0.2in
\end{figure}

In this section, we explain how the proposed region-based energy neural network (RENN) works. In a nutshell, RENN directly minimizes the region-based free energy of a region graph that generalizes the Bethe approximation by amortizing a subset of ${\Bb}$, similar to the Inference Net's direct minimization of Bethe free energy. Please see Appendix~\ref{apdix:sec:get-bethe-from-region-energy} for details on how to recover the Bethe free energy from the region-based free energy. Here, we restrict the beliefs that are directly amortized to be a subset of ${\Bb}$, and recursively compute the remaining beliefs
according to the region graph structure as detailed in Section~\ref{sec:infer-renn}. Then the minimization of region-based free energy is converted into the optimization w.r.t. the parameters of a neural network in RENN. This can reduce the number of neural network parameters compared to directly modeling the beliefs of all regions.

\subsection{Inference by RENN}
\label{sec:infer-renn}

We define some notations that are going to be used in our paper. Since $\Gg_R$ is a hierarchical directed graph, $\Rr_l$ denotes regions in level $l$, and 
$R^{[l]}_i\in \Rr_l$ denotes  the $i$-th region node in level $l$. This means $\Rr_0$ is the set of the top root regions that have no parents  (i.e. the level $0$ regions). Also, $R^{[l]}$ is used to refer to any node in $\Rr_l$, and $R$ denotes a region node when it is not clear or does not matter at which level it is located. Lastly, we define the scope of $R$ by $\Ss(R)$, i.e. $\Ss(R) = \left\{ x_i| i \in R \right\}$.

For a root region $R^{[0]}\!\in\!\Rr_0$, RENN has a corresponding vector representing its score $\bm{f}(\Gg_R, R^{[0]}; \bm{\omega}) \in \RR^{|{\Ss(R^{[0]})}| \times K}$, where $\bm{\omega}$ is the parameter of the mapping $\bm{f}$ that is modeled by a neural network and $|\cdot|$ denotes the cardinality. We define the predicted belief on the root region node $R^{[0]}$ as
\begin{equation}
  b_{R^{[0]}}(\bm{x}_{R^{[0]}}; \bm{\omega}) = \sigma(\bm{f}(\Gg_R, R^{[0]}; \bm{\omega})), \forall~ {R^{[0]}} \in \Rr,
\end{equation}
where $\sigma(\cdot)$ is the softmax function. The softmax function guarantees $b_{R^{[0]}} \in (0,1)^{|{\Ss(R^{[0]})}| \times K}$.

The representation mapping $\bm{f}$ followed by the softmax function in a RENN only needs to directly output the beliefs on root regions in $\Rr_0$, with the dimension of $|\Rr_0| \times |\Ss(R)| \times K$ (assuming the number of variable nodes in each root region is the same).
For the remaining regions $\left\{R \in \Rr \backslash \Rr_0 \right\}$ that are not root regions,
where $\backslash$ denotes the set exclusion, the RENN computes the belief as
\begin{equation}\label{eq:l-level-beliefs}
  b_{R^{[l]}}(\bm{x}_{R^{[l]}}; \bm{\omega}) = \hspace{-0.04cm}\frac{1}{|\Pp(R^{[l]})|}\hspace{-0.0cm} \sum_{R_p\!\in \! \Pp(R^{[l]})}\sum_{\Ss\!(R_p)\backslash \Ss\!(R^{[l]})} \hspace{-0.5cm}b_{R_p}(\bm{x}_{R_p}; \bm{\omega}),
\end{equation}
where $\Pp(R^{[l]})$ is the set of parent regions of $R^{[l]}$ in region graph $\Gg_R$. The non-root region belief of RENN defined in this way comes with the intuition of typical iterative belief propagation methods. In BP and its variants, messages are passed to a variable node to reduce the mismatch of beliefs w.r.t. the variable node, which are sent from this node's neighbors in a factor graph. The message passing iteration of BP or its variants stops when this kind of mismatch w.r.t. every variable node is eliminated in the factor graph.

In RENN, we directly put the mismatch between a non-root region belief $b_{R^{[l]}}(\bm{x}_{R^{[l]}}; \bm{\omega})$ and the marginalization from its parent region $\sum_{\Ss(R_p)\backslash \Ss(R^{[l]})}b_{R_p}(\bm{x}_{R_p}; \bm{\omega})$ as a penalty in the cost function. As the mismatch penalty is close to zero, the non-root region belief gets close to the marginalization calculated from its parent regions. Matching a region's belief with marginalization from its parent regions' beliefs is termed as region belief consistency in region graph.

Different from GBP that minimizes region-based free energy by iterative message-passing, RENN minimizes the region-based free energy by optimizing w.r.t. the neural network parameter $\bm{\omega}$.
Considering the region belief consistency, we summarize the cost function of RENN to include both the region-based free energy and mismatch penalty on non-root regions. This gives the problem
\begin{equation}\label{eq:infer-F-all-belief}
  \umin{\bm{\omega}}{F_R(\Bb;\bm{\theta}) \!+\! \lambda \hspace{-0.3cm}\sum_{R\in \Rr \backslash \Rr_0} \sum_{R_p \in \Pp(R)}\hspace{-0.3cm}d( b_R, \hspace{-0.4cm}\sum_{\Ss(R_p)\backslash \Ss(R)}\hspace{-0.4cm} b_{R_p}(\bm{x}_{R_p}; \bm{\omega}))},
\end{equation}
where $d(\cdot, \cdot)$ is distance metric or divergence to measure the mismatch between the beliefs (the $L_2$ distance is used in our experiments), and $\lambda$ is the regularization parameter.

As shown in Figure~\ref{fig:renn-illustration}, a RENN takes embedding vectors as input and outputs the beliefs on $\Rr_0$ directly (embedding vectors will be explained in Section~\ref{subsec:exp-setting}, although not explicitly included in the objective function \eqref{eq:infer-F-all-belief}). The beliefs in other levels $\{\Rr_1$, $\Rr_2\}$ are computed as in \eqref{eq:l-level-beliefs}. Then the region-based free energy along with the penalty of region belief consistency is minimized w.r.t. $\bm{\omega}$.

\subsection{Region Graph Construction for RENN}
In this section, we explain how to construct the region graph $\Gg_R$ for RENN.
Informally, a region graph can be generated by firstly clustering the nodes in a factor graph in any way and then connecting the clusters with directed edges. Unfortunately, we can not rely on an arbitrary region graph. Conditions such as \textit{valid} region graph (see Section~\ref{subsec:count-number}) and \textit{maxent-normality} \cite{yedida2005constucting,welling2005structured} have been proposed for region graphs, but these conditions do not give rules for how to construct ''good'' region graphs.
We address this issue by combining the cluster variation method \cite{PhysRev.81.988,morita1991cluster} with \textit{tree-robust} condition \cite{gelfand2012generalized} that was originally developed to improve accuracy of GBP, for region graph construction of RENN.

\subsubsection{Determining the Counting Numbers}
\label{subsec:count-number}
In Definition~\ref{def:region-free-energy}, region-based free energy is a function of counting numbers $\left\{ c_R \right\}$. The counting numbers here are used to balance each region's contribution to the free energy. According to \cite{yedida2005constucting}, the region-base free energy is \text{valid} if 
$\sum_{R\in\Rr} c_R \delta_{R}(i)  = 1, \forall~ \mathrm{node}~ i~~ \mathrm{in} ~~\Gg_F$,
where $\delta_{R}(i)$ is the indicator function, equal to $1$ if and only if node $i$ defined in factor graph $\Gg_F$ is in region $R$ of region graph $\Gg_R$, and equal to $0$ otherwise. Note that node $i$ can be either a variable or factor node here. It can be seen that each node would be counted exactly once if the valid condition holds.
Given a region graph $\Gg_R$, the counting numbers $\left\{ c_R \right\}$ can be constructed recursively as:
\begin{equation}
  c_R = 1 - \sum_{R_i \in \Aa(R)} c_{R_i}, \forall R,
\end{equation}
where $\Aa(R)$ denotes the ancestor set of region node $R$ in $\Gg_R$. This rule implies that counting numbers of root regions are always $1$ since they do not have any ancestors.

\subsubsection{Generating Graph by Cluster Variation Method}
\label{sec:cluster-variation-method}
Cluster variation method was introduced by Kikuchi and other physicists \cite{PhysRev.81.988,morita1991cluster}, which started with the intuition of approximating free energy by using larger sets of variable nodes instead of the single-node factorization in the mean field approximation.

The cluster variation method starts with the root regions in $\Rr_0$. There are two requirements for $\Rr_0$: i) every variable node $i$ of factor graph $\Gg_F$ is included in at least one region $R^{[0]}\in\Rr_0$; ii) there should be no region $R^{[0]}\in \Rr_0$ being a subregion of any other region in $\Rr_0$.
With $\Rr_0$ ready, the other sets of regions are generated hierarchically. To construct level-$1$ regions $\Rr_1$ from $\Rr_0$, we find  all the intersections between regions in $\Rr_0$ and omit all that are subregion of other intersection regions. Then level-$2$ regions $\Rr_2$ can be similarly constructed from $\Rr_1$. Assume there are $L$ such sets, then $\Rr = \Rr_0 \cup \Rr_1 \cup \cdots \cup \Rr_{L-1}$. The construction rule can be formulated as
\begin{align}
  \Rr_l = \{ R^{[l]}_i = R^{[l-1]}_j \cap R^{[l-1]}_k | R^{[l]}_i \not\subset R^{[l]}_n,~ \forall i \neq n, R^{[l-1]}_j, R^{[l-1]}_k \in \Rr_{l-1} , j\neq k\},
\end{align}
for $l=1, 2, \cdots, L-1$.
With the hierarchical region sets built, we need to draw the edges. The directed edges are always connected from regions in $\Rr_{l-1}$ to those in $\Rr_{l}$. For one region $R^{[l]}$ in $\Rr_l$, a directed edge is drawn from any superregion of $R^{[l]}$ in $\Rr_l$. This can be represented as
\begin{align}
  \Ee = \{ e = (R^{[l-1]}, R^{[l]}) | R^{[l]} \subset R^{[l-1]}, R^{[l]} \in \Rr_l, R^{[l-1]} \in \Rr_{l-1} , \forall l\}.
\end{align}

\subsubsection{Selection Criteria for Root Regions $\Rr_{0}$}
\label{sec:criteria-root-regions}
Section~\ref{sec:cluster-variation-method} detailed how to construct $\Rr_l$ for $l>0$ from a known $\Rr_0$. We explain how to build the root region set $\Rr_0$ here.

Specifically, we use the \textit{tree-robust} condition \cite{welling2005structured, gelfand2012generalized} to build the root regions for our RENN. We restrict ourselves to construct root regions that are cycles of the factor graph $\Gg_F$. Then constructing root regions for $\Gg_R$ becomes to construct  cycle-structured region sets. A cycle-structured region set becomes a cycle basis when it fulfills certain conditions (see Definition~\ref{apdix:def:cycle-basis} in Appendix~\ref{apdix:sec:root-region-construct}). In a nutshell, the tree-robust condition defines a special class of cycle bases (see Appendix~\ref{apdix:sec:root-region-construct} for the formal definition).
To maintain the consistency, two theorems from \cite{gelfand2012generalized} for choosing tree-robust cycle bases in two graph classes (i.e., planar graphs and complete graphs) are presented here. A planar graph is a graph that can be embedded in the two-dimensional plain (i.e., it can be drawn in the plane such that edges intersect only in their nodes). In a complete graph, every pair of distinct nodes is connected by a unique edge.
\begin{thm}\label{thm:planar-tree-robust}
  In a planar graph $\Gg$, the cycle basis comprised of the faces of the graph $\Gg$ is tree-robust.
\end{thm}
\begin{thm}\label{thm:complete-tree-robust}
  In a complete graph $\Gg$, construct a cycle basis as follows. Choose a node $i$ as the root. Create a 'star' spanning tree rooted at $i$. Then construct cycles of the form $(i,j,k)$ from each off-tree edge $(j,k)$. The constructed basis is tree-robust.
\end{thm}

Tree-robust root regions can also be constructed for general graphs, which is an extension from Theorem~\ref{thm:planar-tree-robust} and \ref{thm:complete-tree-robust}. Please refer to Algorithm~\ref{apdix:alg:root-region-general-graph} in Appendix~\ref{apdix:sec:root-region-construct} for details.

\section{MRF Learning with Inference of RENN}
\label{sec:model-learning-with-renn}
In Section~\ref{sec:infer-renn}, we explained how to do inference with RENN when the parameter $\bm{\theta}$ of $p(\bm{x}; \bm{\theta})$ is assumed to be known. In this section, we  consider the case of learning the parameter $\bm{\theta}$ of an MRF $p(\bm{x}; \bm{\theta})$ with inference by RENN.

When we are given a dataset $\{\bm{x}\}$ and  want to learn the model of $p(\bm{x}; \bm{\omega})$ by maximizing the log-likelihood, it requires to solve
\begin{equation}\label{eq:maximizing-likelihood}
  \umin{\bm{\theta}}{ -\log{\tilde{p}(\bm{x}; \bm{\theta})} + \log{Z(\bm{\theta})}},
\end{equation}
where $\tilde{p}(\bm{x}; \bm{\theta}) =  \prod_{a} \psi_a(\bm{x}_a; \bm{\theta}_a)$. Due to the intractability of $\log{Z(\bm{\theta})}$, it is expensive or prohibitive to solve \eqref{eq:maximizing-likelihood} directly. The minimized region-based free energy $F_R(\Bb;\bm{\theta})$ is exactly the negative log-partition function of $p(\bm{x};\bm{\theta})$, if $b_R(\bm{x}_R)=p(\bm{x}_{R})$, $\forall~R\in \Rr$, as in Section~\ref{sec:preliminary}. We use, $F_R(\Bb^{\ast};\bm{\theta})$ as an approximation to $-Z(\bm{\theta})$ for the general case, where $\Bb^{\ast}= \{b_R(\bm{x}_R; \bm{\omega}^{\ast}), R\in \Rr\}$ with $\bm{\omega}^{\ast}$ being the solution to problem \eqref{eq:infer-F-all-belief}. Combining the MRF learning and RENN inference, we have
\begin{align}\label{eq:learning-min-max}
  \!\!\!\!\!\!\!\!\!\!\min_{\bm{\theta}}\max_{\bm{\omega}} -\log{\tilde{p}(\bm{x}; \bm{\theta})} - F_R(\Bb; \bm{\theta}) -\lambda \!\!\!\!\! \sum_{R\in \Rr \backslash \Rr_0} \sum_{R_p \in \Pp(R)}\!\!\!\!\!d( b_R, \!\!\!\!\! \sum_{\Ss(R_p)\backslash \Ss(R)}\!\!\!\!\! b_{R_p}(\bm{x}_{R_p}; \bm{\omega})).
\end{align}
Then the difficulty of learning of the MRF $p(\bm{x};\bm{\theta})$ is dealt with inference of RENN in \eqref{eq:learning-min-max}.

Note that MRF learning with RENN inference does not rely on sampling to estimate the gradient of the objective. The gradients in \eqref{eq:learning-min-max} can be directly computed with autodiff functions in modern toolboxes such as PyTorch or TensorFlow. Also, since RENN does not need iterative message propagation, MRF learning with inference by RENN can be faster. Finally, our method can be extended to learn models where there are both observable variable $\bm{x}$ and hidden variable $\bm{z}$ that we do not have observations for. Please refer to Appendix~\ref{apdix:sec:crf-learning} for further discussions.

\section{Experimental Results}

We conducted a series of experiments to validate the proposed RENN model, in both inference and learning problems of MRFs. The experiment code is attached to the submission. Code is available at https://github.com/FirstHandScientist/renn.

\subsection{Experiment Setting and Evaluation Metrics}
\label{subsec:exp-setting}
Without loss of generality, our experiments are carried out on binary pairwise MRF (Ising model). This gives us $p(\bm{x}; \bm{\theta}) = \frac{1}{Z(\bm{\theta})}\exp{(\sum_{(i,j)\in \Ee_F} J_{ij} x_i x_j + \sum_{i\in \Vv}h_i x_i)}$, $\bm{x} \in \{-1, 1\}^{N}$, where $J_{ij}$ is the pairwise log-potential between node $i$ and $j$, $h_i$ is the node log-potential for node $i$. Then $\bm{\theta} = \left\{ J_{ij}, h_i| (i,j) \in \Ee_F, i \in \Vv \right\}$. $J_{ij}$ is always sampled from standard normal distribution, i.e. $J_{ij}\sim \Nn(0,1)$; meanwhile $h_i \sim \Nn(0, \gamma^{2})$ with $\gamma$ reflecting the relative strength of univariate log-potentials to pairwise log-potentials

In the inference experiments, we are interested in how well beliefs from RENN approximate true marginal distributions of $p(\bm{x};\bm{\theta})$. We quantify this by both the $\ell_1$ error (i.e., $\ell_1$-norm distance) and the Pearson correlation coefficient $\rho$ between the true marginals and beliefs of RENN. The evaluations include both $p(x_i)$ and $p(x_i,x_j)$ and compare with true marginals. Thus, the $\ell_1$ error reflects both the inference error as well as the belief consistency since true marginals are definitely consistent (see Appendix~\ref{apdx:sec:discussion-lambda} for further discussion). In addition, we also quantify the $\log{Z}$ error as the absolute difference between true negative log-partition function and free energy of each approximation method.

Apart from the inference experiments, we also carried out MRF learning experiments. We use the negative log-likelihood (NLL) to evaluate how well an MRF is learned from random parameter initialization, which is then compared with the MRF with true parameterization.

In all experiments and for each evaluation of RENN, mean field, (loopy) BP \cite{mooij2007sufficient}, damped BP \cite{Pretti2005damping} with damping factor $0.5$, and GBP \cite{yedida2005constucting} are  evaluated as benchmarks on the same MRF and compared with RENN. The hyperparameter $\lambda$, regulating the belief consistency (see Appendix~\ref{apdx:sec:discussion-lambda}), is  selected from $\left\{ 1,3,5,10 \right\}$. The neural network benchmark model saddle-point Inference Net \cite{NIPS2019_9687} targeting the Bethe free energy, is also used for comparison. To make the comparison with Inference Net fair, RENN and Inference Net use the same neural network structures and hidden dimension. Each variable $x_i$ is associated with a learnable embedding vector $\bm{e}_i$. A transform layer \cite{AshishNIPS2017_7181} consumes $\bm{e}_i$ and outputs a hidden representation $\bm{h}_i$. The transform layer is shared by all embeddings. Then an affine layer followed by softmax consumes $[\bm{h}_1, \cdots, \bm{h}_N]$ and outputs the beliefs.

\subsection{Inference on Grid Graphs}
\label{sec:inference-grid}
We first evaluate how well RENN can estimate the marginal distributions compared with benchmark algorithms/models w.r.t. marginal $\ell_1$ errors and Pearson correlation $\rho$ for different graph sizes $n$ and standard deviations $\gamma$ of $\{h_i\}$. At each evaluation for a given size $n$ and $\gamma$, $20$ MRFs are generated by sampling $\{J_{ij}\}$ and $\{h_i\}$. Then RENN and other candidate algorithms perform inference on these MRFs. The $\ell_1$ error and correlation $\rho$ between true and estimated marginal distributions are evaluated. The $\log{Z}$ errors are also recorded. The results are reported as 'mean $\pm$ standard deviation'. Partial results are presented here, and results for richer settings are reported in Appendix~\ref{apdx:sec:extra-grid-results}.

% \begin{table}[h]
%   \vskip -0.2in
%   \caption{Inference with the \textit{infinite face} on grid, $n=25$.}
%   \label{tab:infer-infinite-face}
%   \begin{center}
%     \begin{small}
%       \begin{sc}
%         \begin{tabular}{llcc}
%           \toprule
%           $\gamma$ & Metric & GBP & RENN \\
%           \midrule
%           \multirow{3}{*}{0.1}
%                    & $\ell_1$ Error & 0.061 $\pm$ 0.025 & \textbf{0.025} $\pm$ 0.020 \\

%                    & $\rho$   & 0.913 $\pm$ 0.049  &  \textbf{0.984} $\pm$ 0.021  \\
%                    & $\log{Z}$ Error & 3.564 $\pm$ 2.823  &  0.384 $\pm$ 0.223  \\
%           \midrule
%           \multirow{3}{*}{1}
%                    & $\ell_1$ Error & 0.145 $\pm$ 0.028  & \textbf{0.016} $\pm$ 0.010 \\

%                    & $\rho$   & 0.783 $\pm$ 0.091  &  \textbf{0.995} $\pm$ 0.010 \\
%                    & $\log{Z}$ Error & 0.825 $\pm$ 0.841  & 0.364 $\pm$ 0.201 \\
          
%           \bottomrule
%         \end{tabular}
%       \end{sc}
%     \end{tiny}
%   \end{center}
%   \vskip -0.2in
% \end{table}

\begin{table*}[t]
  \caption{Inference on grid graph ($\gamma=0.1$). $\ell_1$ error and correlation $\rho$ (evaluation based on both univariate and pairwise marginals, i.e., $p(x_i)$ and $p(x_i,x_j)$), and $\log{Z}$ error.}
  \label{table:infer-grid-gamma0.1}
  \vskip -0.05in
  \centering
  \begin{adjustbox}{width=1\textwidth}
    \begin{small}
        \begin{tabular}{l c ccccccc}
          \toprule
           Metric & $n$ & Mean Field & Loopy BP & Damped BP & GBP & Inference Net & RENN \\
          \midrule
          \multirow{2}{*}{\begin{tabular}[x]{@{}c@{}}$\ell_1$\\error \end{tabular} }
                 % &    25   &$0.271 \pm 0.051$ &  $0.086 \pm 0.078$ & $0.084 \pm 0.076$ & $0.057 \pm 0.024$ & $0.111 \pm 0.072$ & \textbf{0.049} $\pm$ 0.078 \\
          
                 &    100   & $0.283 \pm 0.024$ &  $0.085 \pm 0.041$ & $0.062 \pm 0.024$ & $0.064 \pm 0.019$ & $0.074 \pm 0.034$ & \textbf{0.025} $\pm$ 0.011 \\
          
                 % &    225   & $0.284 \pm 0.019$ &  $0.100 \pm 0.025$ & $0.076 \pm 0.025$ & $0.073 \pm 0.013$ & $ 0.073 \pm 0.012$ & \textbf{0.046} $\pm$ 0.011 \\
          
                 &    400   & $0.279 \pm 0.014$ &  $0.110 \pm 0.016$ & $0.090 \pm 0.016$ & $0.079 \pm 0.009$ & $ 0.083 \pm 0.009$ & \textbf{0.061} $\pm$ 0.009 \\

          \midrule
          \multirow{2}{*}{\begin{tabular}[x]{@{}c@{}}Correl-\\ation $\rho$ \end{tabular}}
                 % &   25    & 0.633 $\pm$ 0.197  &  0.903 $\pm$ 0.114  &  0.905 $\pm$ 0.113  &  0.923 $\pm$ 0.045  &  0.866$\pm$ 0.117 &  \textbf{0.951} $\pm$ 0.112 \\
          
                 &   100   & 0.582 $\pm$ 0.112  &  0.827 $\pm$ 0.134  &  0.902 $\pm$ 0.059  &  0.899 $\pm$ 0.043  &  0.903$\pm$ 0.049 &   \textbf{0.983} $\pm$ 0.012 \\
          
                 % &   225   & 0.580 $\pm$ 0.080  &  0.801 $\pm$ 0.078  &  0.863 $\pm$ 0.088  &  0.869 $\pm$ 0.037  & 0.873 $\pm$ 0.037 &  \textbf{0.949} $\pm$ 0.022 \\
          
                 &   400   & 0.596 $\pm$ 0.054  &  0.779 $\pm$ 0.059  &  0.822 $\pm$ 0.047  &  0.852 $\pm$ 0.024  & 0.841 $\pm$ 0.028 &  \textbf{0.912} $\pm$ 0.025 \\

          \midrule
          \multirow{2}{*}{\begin{tabular}[x]{@{}c@{}}$\log{Z}$ \\error\end{tabular}}
                 % &   25    & 2.512 $\pm$ 1.060  &  0.549 $\pm$ 0.373  &  0.557 $\pm$ 0.369  &  \textbf{0.169} $\pm$ 0.142  &  0.762 $\pm$ 0.439  &  0.240 $\pm$ 0.140 \\

                 &  100    & 13.09 $\pm$ 2.156  &  1.650 $\pm$ 1.414  &  1.457 $\pm$  1.365 &  \textbf{0.524} $\pm$ 0.313  &  2.836 $\pm$ 2.158  & 1.899 $\pm$ 0.495 \\

                 % &  225    & 29.93 $\pm$ 4.679  &  3.348 $\pm$ 1.954  &  3.423 $\pm$ 2.157  &  \textbf{1.008} $\pm$ 0.653  &  3.249 $\pm$ 2.058  & 4.344 $\pm$ 0.813  \\

                 &  400    & 51.81 $\pm$ 4.706  &  5.738 $\pm$ 2.107  &  5.873$\pm$ 2.211   &  \textbf{1.750} $\pm$ 0.869  &  3.953 $\pm$ 2.558  & 7.598 $\pm$ 1.146 \\
          
                 % \midrule
                 % \multirow{4}{*}{Time}
                 % &   25   &  0.259 $\pm$ 0.076  &  2.990 $\pm$ 4.563  &  1.591 $\pm$ 0.609  &  6.817 $\pm$ 0.339  &  5.253 $\pm$ 1.189  &  5.828 $\pm$ 2.372  \\
                 % &  100   &  1.290 $\pm$ 0.205  &  57.10 $\pm$ 25.64 &  28.70 $\pm$ 24.62 & 46.89$ \pm$ 1.365 & 22.05 $\pm$ 10.53  & 22.63 $\pm$ 6.202 \\

                 % &  225   &  3.838 $\pm$ 1.015  &  184.1 $\pm$ 3.648  &  124.2 $\pm$ 68.70  &  131.1 $\pm$ 3.440  &  36.43 $\pm$ 2.087  & 51.82 $\pm$ 2.431 \\

                 % &  400   &  7.908 $\pm$ 1.508  & 335.3 $\pm$ 12.83   &  326.9 $\pm$ 27.56  &  253.7 $\pm$ 11.00   &  69.54 $\pm$ 5.140  & 103.9 $\pm$ 8.379 \\

          \bottomrule
        \end{tabular}
  \end{small}
  \end{adjustbox}
    \vskip -0.17in
\end{table*}

\begin{table*}[t!]
  \caption{Inference on complete graph of size $16$.}
  \label{tab:infer-full-n16}
  \vskip -0.05in
  \centering
  \begin{adjustbox}{width=1\textwidth}
    \begin{small}
        \begin{tabular}{l p{0.01cm} p{1.8cm} p{1.8cm} p{1.8cm} p{1.8cm} p{1.8cm} p{1.8cm}}
          \toprule
          Metric & $\gamma$ & Mean Field & Loopy BP & Damped BP & GBP & Inference Net & RENN \\
          \midrule
          \multirow{2}{*}{\begin{tabular}[x]{@{}c@{}}$\ell_1$-\\error\end{tabular}}
          % pen3
                 % & 0.1   &  0.303 $\pm$ 0.056  &  0.176 $\pm$ 0.039  &  0.174 $\pm$ 0.038  &  0.244 $\pm$ 0.047  &  0.174 $\pm$ 0.044  &  \textbf{0.169} $\pm$ 0.052  \\

                 % pen2
                 &  1    &  0.273 $\pm$ 0.086  &  0.239 $\pm$ 0.059  &  0.239 $\pm$ 0.059  &  0.260 $\pm$ 0.086  &  0.249 $\pm$ 0.067  &  \textbf{0.181} $\pm$ 0.092  \\
          
                 % pen2
                 % &  2    &  0.231 $\pm$ 0.079  &  0.222 $\pm$ 0.064  &  0.221 $\pm$ 0.064  &  0.249 $\pm$ 0.078  &  0.232 $\pm$ 0.069  &  \textbf{0.170} $\pm$ 0.109  \\

                 % % pen3
                 % &  3    &  0.218 $\pm$ 0.042  &  0.204 $\pm$ 0.038  &  0.204 $\pm$ 0.038  &  0.247 $\pm$ 0.065  &  0.213 $\pm$ 0.051  &  \textbf{0.138} $\pm$ 0.106  \\

                 % pen1
                 &  4    &  0.197 $\pm$0.049   &  0.181 $\pm$ 0.035  &  0.180 $\pm$ 0.034  &  0.210 $\pm$ 0.070  &  0.174 $\pm$ 0.030  &  \textbf{0.125} $\pm$ 0.050  \\

          \midrule
          \multirow{2}{*}{\begin{tabular}[x]{@{}c@{}}Correl-\\ation $\rho$\end{tabular}}
                 % & 0.1   &  0.231 $\pm$ 0.196  &  0.509 $\pm$ 0.056  &  0.510 $\pm$ 0.055  &  0.316 $\pm$ 0.207  &  0.506 $\pm$ 0.063  &  \textbf{0.539} $\pm$ 0.235  \\

                 &  1    &  0.381 $\pm$ 0.255  &  0.514 $\pm$ 0.185  &  0.515 $\pm$ 0.185  &  0.445 $\pm$ 0.223  &  0.533 $\pm$ 0.150  &  \textbf{0.756} $\pm$ 0.187  \\

                 % &  2    &  0.535 $\pm$ 0.207  &  0.569 $\pm$ 0.180  &  0.570 $\pm$ 0.179  &  0.480 $\pm$ 0.186  &  0.559 $\pm$ 0.176  &  \textbf{0.750} $\pm$ 0.261  \\

                 % &  3    &  0.586 $\pm$ 0.142  &  0.618 $\pm$ 0.134  &  0.619 $\pm$ 0.134  &  0.502 $\pm$ 0.144  &  0.613 $\pm$ 0.128  &  \textbf{0.853} $\pm$ 0.159  \\

                 &  4    &  0.622 $\pm$ 0.166  &  0.658 $\pm$ 0.133  &  0.660 $\pm$ 0.132  &  0.564 $\pm$ 0.165  &  0.693 $\pm$ 0.060  &  \textbf{0.868} $\pm$ 0.053  \\

          \midrule
          \multirow{2}{*}{\begin{tabular}[x]{@{}c@{}}$\log{Z}$ \\error\end{tabular}}
                 % & 0.1   &  24.45 $\pm$ 7.560  &  143.7 $\pm$ 9.297  &  145.5 $\pm$ 6.096  &  166.3 $\pm$ 11.98  &  148.5 $\pm$ 3.522  &  \textbf{12.57} $\pm$ 3.689  \\

                 &  1    &  20.66 $\pm$ 5.451  &  178.7 $\pm$ 22.18  &  178.9 $\pm$ 21.88  &  153.3 $\pm$ 25.29  &  213.6 $\pm$ 12.75  &  \textbf{14.41} $\pm$ 4.135  \\

                 % &  2    &  16.04 $\pm$ 4.352  &  296.3 $\pm$ 44.41  &  296.9 $\pm$ 44.24  &  116.9 $\pm$ 32.72  &  335.1 $\pm$ 32.86  &  \textbf{13.37} $\pm$ 4.531  \\

                 % &  3    &  13.87 $\pm$ 6.554  &  432.7 $\pm$ 66.44  &  433.4 $\pm$ 66.30  &  100.2 $\pm$ 39.62  &  462.9 $\pm$ 53.61  &  \textbf{12.56} $\pm$ 6.046  \\

                 &  4    &  \textbf{10.74} $\pm$ 7.385  &  565.7 $\pm$ 73.33  &  566.1 $\pm$ 73.13  &  106.0 $\pm$ 54.43  &  588.3 $\pm$ 62.58  &  14.72 $\pm$ 4.155  \\

                 % \midrule
                 % \multirow{4}{*}{Time}
                 % & 0.1   &  0.555 $\pm$ 0.135  &  1.705 $\pm$ 1.256  &  1.428 $\pm$ 0.400  &  13.68 $\pm$ 0.432  &  12.63 $\pm$ 0.592  &  15.26 $\pm$ 4.759  \\

                 % &  1    &  0.457 $\pm$ 0.103  &  0.777 $\pm$ 0.167  &  1.285 $\pm$ 0.190  &  14.29 $\pm$ 0.487  &  12.45 $\pm$ 0.603  &  16.16 $\pm$ 5.695  \\

                 % &  2    &  0.424 $\pm$ 0.101  &  0.630 $\pm$ 0.096  &  1.143 $\pm$ 0.196  &  14.76 $\pm$ 0.512  &  15.04 $\pm$ 1.106  &  14.74 $\pm$ 4.606  \\

                 % &  3    &  0.377 $\pm$ 0.114  &  0.567 $\pm$ 0.118  &  1.084 $\pm$ 0.254  &  14.49 $\pm$ 0.392  &  15.62 $\pm$ 1.218  &  13.40 $\pm$ 3.070  \\

                 % &  4    &  0.521 $\pm$ 0.475  &  0.543 $\pm$ 0.137  &  1.018 $\pm$ 0.171  &  14.00 $\pm$ 0.379  &  14.59 $\pm$ 5.254  &  14.24 $\pm$ 2.536 \\

          \bottomrule
        \end{tabular}
      \end{small}
      \end{adjustbox}
    \vskip -0.2in
\end{table*}

The $\ell_1$ error and correlation coefficient $\rho$ reflect the marginal approximation quality directly and indicate the consistency of the beliefs (except for the mean field method). The results are reported in Table~\ref{table:infer-grid-gamma0.1}, and additional results are included in  Appendix~\ref{apdx:sec:extra-grid-results}. Beliefs of RENN outperform benchmark algorithms for marginal inference. As expected, the performance of loopy BP and its variant damped BP are similar in general while damped BP sometimes gets better estimations. Both loopy BP and damped BP have better marginal estimations than the mean field method in all of our considered scenarios. GBP outperforms loopy BP and damped BP for $\gamma=0.1$, agreeing with the results from \cite{yedida2005constucting}, but performs poorly for $\gamma=1$ in Appendix~\ref{apdx:sec:extra-grid-results}. Similar phenomena can be observed for Inference Net. As for the error of the partition function values, GBP gets the most accurate estimations when $\gamma=0.1$. Partition function estimation by RENN is competitive in the different considered cases.

Note the region graphs in this set of experiments use all faces of a grid graph but the \textit{infinite face} (the perimeter circle). The performance of RENN can be further improved by including the infinite face (see Table~\ref{apdx:tab:infer-infinite-face} in Appendix~\ref{apdx:sec:extra-grid-results}).

\begin{table*}[t]
  \caption{NLL of MRF learning using different inference methods.}
  \label{tab:nll-training-grid-complete}
  \vskip -0.05in
  \centering
    \begin{small}
      \bgroup
      \def\arraystretch{0.9}
        \begin{tabular}{lcccccccc}
          % std=1.0
          \toprule
          $n$ & True & Exact & Mean Field & Loopy BP & Damped BP & GBP & Inference Net & RENN \\
          \toprule
          \multicolumn{9}{c}{Grid Graph}\\
          \midrule
          25  &  9.000  &  9.004  &  9.811  &  {9.139}  &  9.196  &  10.56  &  9.252  &  \textbf{9.048}  \\
          100 &  19.34  &  19.38  &  23.48  &  {19.92}  &  20.02  &  28.61  &  20. 29  &  \textbf{19.76} \\
          225 &  63.90  &  63.97  &  69.01  &  66.44    &  66.25  &  92.62  &  68.15  &  \textbf{64.79}  \\
          \toprule
          % std=1.0
          \multicolumn{9}{c}{Complete Graph}\\
          \midrule
          9  &  3.276  &  3.286  &  9.558  &  5.201  &  5.880  &  10.06  &  5.262  & \textbf{3.414}  \\
          16  &  4.883  &  4.934  &  28.74  &  13.64  &  18.95  &  24.45  &  13.77  &  \textbf{5.178}  \\

          \bottomrule
        \end{tabular}
        \egroup
      \end{small}
  \vskip -0.22in
\end{table*}

% \begin{table}[t]
%   \caption{Consumed time per epoch (unit second).}
%   \label{tab:time-training}
%   \begin{center}
%     \begin{small}
%       \begin{sc}
%         \begin{tabular}{lcc}
%           \toprule
%           $n$ & 25 & 100 \\
%           \midrule
%           Mean Field & 8.850 & 24.36 \\
%           Loopy BP &  41.58 & 94.97 \\
%           Damped BP & 35.85 & 156.8 \\
%           GBP &  1.466 & 9.245  \\
%           Inference Net & 1.466 & 5.314 \\
%           RENN &  2.329 & 10.98\\

%           \bottomrule
%         \end{tabular}
%       \end{sc}
%     \end{small}
%   \end{center}
%   \vskip -0.2in
% \end{table}

\begin{table}[t!]
  \caption{Average consumed time per epoch (unit: second) for two learning cases in Table~\ref{tab:nll-training-grid-complete}.}
  \label{tab:time-training}
  \centering
  \begin{small}
        \begin{tabular}{lcccccc}
          \toprule
          {} &  Mean Field & Loopy BP & Damped BP & GBP & Inference Net & RENN \\
          \toprule
          Grid $\Gg$, $n\!=\!225\!\!$ & 40.09 & 335.1 & 525.1 & 12.37 & 19.49 & 16.03 \\
           Complete $\Gg$, $n=\!16\!\!$ & 2.499 & 12.40 & 5.431 & 1.387 & 0.882 & 2.262 \\
          % \midrule
        %   25  &  9.000  &  9.004  &  9.811  &  {9.139}  &  9.196  &  10.56  &  9.252  &  \textbf{9.048}  \\

        % \begin{tabular}{lcc}
        %   \toprule
        %   {} & \begin{tabular}[x]{@{}c@{}} Grid Graph\\ $n=225$\end{tabular}
        %      & \begin{tabular}[x]{@{}c@{}} Complete Graph\\ $n=16$\end{tabular}  \\
        %   \midrule
        %   Mean Field & 40.09 & 2.499 \\
        %   Loopy BP &  335.1 & 12.40\\
        %   Damped BP & 525.1 & 5.431\\
        %   GBP &   12.37    & 1.387\\
        %   Inference Net & 19.49 & 0.882 \\
        %   RENN & 16.03  & 2.262\\

          \bottomrule
        \end{tabular}
    \end{small}
  \vskip -0.2in
\end{table}

\subsection{Inference on Challenging Complete Graphs}
\label{apdx:sec:infer-complete-graph}
In this section, we compare RENN with benchmark methods on more challenging complete graphs, in which every two nodes are connected by a unique edge. Due to the high complexity, we carry out the inference experiments on complete graphs of size $n=16$ but with a richer setting of $\gamma$ (see Appendix~\ref{apdx:sec:extra-result-complete} for more results of different graph sizes and $\gamma$ configurations), to be able to track the true marginals and partition functions exactly, which are used to evaluate candidate methods.

In this comparison, RENN still outperforms almost all other benchmark methods in both marginal and partition function estimation. Different from the case of grids, the benchmark methods except for the mean field, return large errors for the partition function estimates, which may due to convergence issues in challenging complete graphs. RENN still gives competitive results. Similar phenomena could be observed for a different setting of relative potential strength and graph sizes (see Appendix~\ref{apdx:sec:extra-result-complete}).

\subsection{MRF Learning with Inference of RENN}

In this section, we report the results of learning MRFs, i.e. learning the MRF parameter $\bm{\theta}$ as discussed in Section~\ref{sec:model-learning-with-renn}, by inference of RENN.

We do MRF learning on two types of graphs. For both cases, we firstly sample the parameter set $\bm{\theta}^{\prime}$, and then sample training and testing dataset from $p(\bm{x}; \bm{\theta}^{\prime})$. The true NLL of the sampled datasets can be computed by $p(\bm{x}; \bm{\theta}^{\prime})$. We then do learning that starts from a randomly-initialized MRF with the obtained training dataset by inference of RENN (see Section~\ref{sec:model-learning-with-renn}). The learned MRF by RENN is evaluated with the testing dataset w.r.t. the NLL value, which is compared with learned MRFs by other methods. We also include the comparison with exact inference where $Z(\bm{\theta})$ is computed exactly.
In the grid graphs, there are $4000$ samples for MRF learning and $1000$ for testing. In the complete graph case, there are $2000$ samples for MRF learning and $1000$ samples for testing.

In the cases of both grid and complete graphs, RENN shows advantageous performance as shown in Table~\ref{tab:nll-training-grid-complete} with larger marginal in challenging complete graphs. Additionally, RENN is much faster compared with message passing algorithms. As shown in Table~\ref{tab:time-training}, loopy BP needs almost $335$s and damped BP needs about $525$s per epoch iteration, while RENN takes $16$s per epoch. Please refer to Appendix~\ref{apdx:sec:extrain-learning-time} for computation time of other cases.
Neural network based methods parameterize the beliefs or marginal distributions and thus can do new inference estimations much faster when model parameter $\bm{\theta}$ is updated in optimization steps.

\section{Related Work}

Neural networks are popularly used in deep graphical generative models for structured data modeling \cite{qu2019gmnn, johansonNIPS2016_6379, li2018graphical}. Along with a neural network based generative model, a separate neural network has to be trained for inference or recognition. In these directed graphical models built on neural networks, training of inference networks needs sampling which brings in the trade-off between training speed and estimation variance. These issues also lies in the VAE \cite{DBLP:journals/corr/KingmaW13,2017arXiv170104722M}, NVIL \cite{kuleshov2017neural_variational}, AdVIL \cite{li2019AdVIL} and other variational methods \cite{NIPS2017_7136}.

Apart from the directed graphical models, there is also a track of work on using neural networks to model the message passing functions. \cite{akbayrak2019reparameterization} models the intractable message update functions by a Gaussian distribution with its parameters as the output of a neural network, and then follows the typical message passing rules to do iterative message updates of standard BP. \cite{jitkrittum2015kernel, heess2013learning} also similarly learn a neural network to model the message update functions of expectation propagation methods.

Note that although recent neural message passing methods are also purely neural network based models for inference tasks, these methods still do iterative message propagation analogous to standard BP.
Neural message passing methods \cite{yoon2019inferenceGraph, pmlr-v70-gilmer17a} use a  graph network update messages and a separate network to map messages into targeted results. Training of these models has to rely on sampling methods since true messages or marginals are usually not available.

\section{Conclusion}
We presented a neural network based model, RENN, to do inference in MRFs and also learning of MRFs. The proposed model is verified via experiments and is shown to outperform the benchmark methods. It would be interesting to investigate the applications of RENN to variants of MRFs in future work.

% \section*{Broader Impact}
% The broader impact does not apply to our work since our work studies the fundamental inference problems in probabilistic graphical models.

\small

\bibliography{myref}
\bibliographystyle{plain}
\newpage
\appendix
\section{Variational Free Energy and Mean Field}
\label{apdx:sec:variational-free-energy-and-mf}
Variational approaches essentially use a simple analytic form for approximation to the true distribution. It starts from \textit{variational free energy} \cite{opper2001advanced}, where a probability distribution $b(\bm{x})$ is used to approximate $p(\bm{x};\bm{\theta})$ (defined in \eqref{eq:joint-px} in our paper). The variational free energy is defined by
\begin{align}
  F_V(b) &= \sum_{\bm{x}}b(\bm{x}) \ln{\frac{b(\bm{x})}{{p}(\bm{x}; \bm{\theta})}} - \ln{Z(\bm{\theta})} \nonumber \\
         & = \sum_{\bm{x}}b(\bm{x}) \ln{\frac{b(\bm{x})}{\tilde{p}(\bm{x};
           \bm{\theta})}} \nonumber \\
         & = \mathrm{KL}(b( \bm{x}) || \tilde{p}(\bm{x}; \bm{\theta}))
\end{align}

where $\tilde{p}(\bm{x}; \bm{\theta}) =  \prod_{a} \psi_a(\bm{x}_a; \bm{\theta}_a)$ and $\mathrm{KL}(\cdot || \cdot)$ is the Kullback-Leibler divergence.

In the mean field approach, a fully-factorized approximation is used, i.e., $b(\bm{x})$ is a fully-factorized probability distribution with the form
\begin{equation}\label{eq:mf-factorization}
  b_{MF}(\bm{x}) = \prod_{i=1}^{N}b_i(x_i).
\end{equation}
Substituting \eqref{eq:mf-factorization} into the variational free energy gives
\begin{align}
  F_{MF} =  - \sum_{a\in\Ff}\sum_{\bm{x}_a} \ln{\psi_a(\bm{x}_{a};\bm{\theta})} \prod_{i\in \mathrm{ne}_a}b_i(x_i)  + \sum_{i=1}^{N} \sum_{x_i} b_i(x_i) \ln{b_i(x_i)},
\end{align}
where $\mathrm{ne}_a = \left\{ i \in \Vv | x_i \in \Ss(a) \right\}$ denotes the neighboring variable nodes of the factor node $a$, and $\Ss(a)$ is the scope set (arguments) of factor node $a$ as defined in our paper.
Solving the minimization of $F_{MF}$ w.r.t. $b_{MF}(\bm{x})$ gives the
update rule of mean field as
\begin{equation}\label{apdx:eq:mf-update}
  \ln{b_i(x_i)} \propto \sum_{a\in \mathrm{ne}_i} \sum_{\bm{x}_a \backslash x_i} \ln{\psi_a}(\bm{x}_a;\bm{\theta}_a) \prod_{j\in \mathrm{ne}_a\backslash i} b_j(x_j),
\end{equation}
where $\mathrm{ne}_i = \left\{ a | i \in \Ss(a), a \in \Ff \right\}$, i.e. the neighboring factors of node $i$, $\propto$ stands for 'proportional to'. The right-hand-side of $\propto$ is a function of neighboring potential functions and beliefs of node $i$, which computes the new belief $b_i(x_i)$. Essentially, the variable nodes in $\Vv$ take turns to get updated by following the update rule in \eqref{apdx:eq:mf-update} until convergence or a stop condition is fulfilled.

\section{Bethe Free Energy and (Loopy) Belief Propagation}
\label{apdix:sec:bethe-lBP}

Different from the mean field approximation, Bethe approximation also includes the multivariate beliefs $\{b_a(\bm{x}_a)\}$ apart from the univariate beliefs $\{b_i(x_i)\}$ \cite{yedidia2003understanding}. In this case, the Bethe free energy is given by \eqref{eq:bethe-free-energy} in our paper, which is a function of $\{b_i(x_i), b_a(\bm{x}_a)\}$.
Due to multivariate beliefs, there are consistency constrains $\sum_{\bm{x}_a} b_a(\bm{x}_a) = \sum_{ x_i} b_i({x}_i) =1$, $\forall~ i \in \Ss(a)$ to obey, which makes the problem different from the mean field approximation. Then, the Bethe free energy minimization problem can be formulated as
\begin{align}\label{apdix:eq:bethe-min}
  \min_{\{b_a(\bm{x}_a)\}, \{b_i(x_i)\}}& F_{B} \nonumber \\
  \mathrm{s.t.}~~ & \sum_{\bm{x}_a \backslash x_i} b_a(\bm{x}_a)  =
                    b_i(x_i), \nonumber \\
                                        & \sum_{\bm{x}_a} b_a(\bm{x}_a) = \sum_{ x_i} b_i({x}_i) =1,
                                          \nonumber \\
                                        &  0 \leq b_i(x_i) \leq 1,  \nonumber \\
                                        &  b_a(\bm{x}_a) \in [0,1]^{|S(\bm{x}_a)|\times K}, \nonumber \\
                                        & i \in \Vv , a \in \Ff,
\end{align}
where $\Vv$ and $\Ff$ are the set of variable nodes and the set of
factor nodes in factor graph as defined in
Definition~\ref{def:factor-graph} in our paper. Solving the Bethe free energy minimization problem \eqref{apdix:eq:bethe-min} gives the message-passing rule
\begin{equation}\label{apdix:eq:lbp-update-rule}
  m_{a\rightarrow i}(x_i) \propto \sum_{\bm{x}_a \backslash x_i}
  \psi_a(\bm{x}_a) \prod_{j \in \Ss(a) \backslash i} \prod_{b \in \mathrm{ne}_j
    \backslash a} m_{b\rightarrow j}(x_j),
\end{equation}
which is know as (loopy) BP. In loopy BP, the message propagation and update under the rule \eqref{apdix:eq:lbp-update-rule} until a stop criteria is meet or convergence. Intuitively, the message passing phase of loopy BP can be viewed as a process of minimizing the Bethe free energy.

\section{Recover Bethe Free Energy from Region-based Free Energy}
\label{apdix:sec:get-bethe-from-region-energy}
Region-based free energy is known to generalize the Bethe free energy. In other words, the Bethe free energy defined in \eqref{eq:bethe-free-energy} can be directly constructed
from the Definition~\ref{def:region-free-energy} in our paper, with
a specific choice of regions. As shown in \cite{yedida2005constucting},
if we define two types of regions (large regions and small
regions) directly from a factor graph $\Gg_F(\Vv \cup \Ff, \Ee_F)$ by defining the large regions and small regions as
\begin{align}
  \Rr_{L} &= \left\{ \left\{ a, \Ss(a) \right\}| a\in \Ff \right\}, \nonumber \\
  \Rr_{S} &= \left\{ \left\{ i \right\} | i \in \Vv \right\}.
\end{align}
It can be seen that these regions fulfill the Definition~\ref{def:region-graph} in our paper. According to Section~\ref{subsec:count-number} in the paper, the large regions always have counting number $c_{R,a}=1$ and for small regions each node $i$ always has counting number $c_{R,i}=1-|\mathrm{ne}_i|$. Then we can recover the Bethe free energy from region-based free energy defined in
\eqref{eq:def-region-free-energy} in our paper. To be specific, for
large regions,
\begin{align}\label{apdix:eq:large-region}
  F_{R,L}(\Bb; \bm{\theta}) &=\sum_{R\in \Rr_{L}} c_{R,a}
                              \sum_{\bm{x}_a}b_a(\bm{x}_a) (E_a(\bm{x}_a) + \ln{b_a}(\bm{x}_a))
                              \nonumber \\
                            & =\sum_{a} \sum_{\bm{x}_a} b_a(\bm{x}_a)\ln{\frac{b_a(\bm{x}_a)}{\psi_a(\bm{x}_a)}}.
\end{align}
And for the small regions, the free energy can be similarly obtained as
\begin{equation}\label{apdix:eq:small-region}
  F_{R,S}  =  \sum_{i=1}^{N} (1- |\mathrm{ne}_i|) \sum_{x_i} b_i(x_i) \ln{b_i(x_i)}.
\end{equation}
Putting \eqref{apdix:eq:large-region} and \eqref{apdix:eq:small-region} together gives the Bethe free energy in \eqref{eq:bethe-free-energy} in the paper.

\section{Generalize Belief Propagation}
\label{apdix:sec:gbp}
The region graph was original proposed for the generalized belief propagation (GBP) message-passing algorithm\cite{Yedidia:2000:GBP:3008751.3008848,yedida2005constucting,DBLP:journals/corr/abs-1207-4158}. We give the message-passing rules of GBP here since it is used as a benchmark comparison method in our paper.

GBP operates on a directed region graph $\Gg_R(\Rr,\Ee)$. A message is always sent from a parent region $P$ to a child region $R$, i.e. over a directed edge $(P, R)\in \Ee$. Let us define the factors in region $R$ as $A_R = \left\{ a | a \in R \right\}$.
Similar to the notation in the paper, $\Pp(R)$ denotes the set of parent regions of $R$. The descendants of $R$ is denoted by $\Dd(R)$ (excluding $R$). The descendants of $R$ including $R$ is denoted by $\hat{\Dd}(R) = \Dd(R) \cup R$. The message update rule from the parent region $P$ to the child region $R$ is
\begin{align}\label{apdix:gbp-update-rule}
  m_{P\rightarrow R} \propto \frac{\sum_{\Ss(P)\backslash
  \Ss(R)}\prod_{a\in A_P\backslash A_R}\psi_a(\bm{x}_a)\prod_{(I,J)
  \in \Nn(P, R)}m_{I\rightarrow J}(\bm{x}_J)}{\prod_{(I,J)
  \in \Hh(P, R)}m_{I\rightarrow J}(\bm{x}_J)},
\end{align}
where
\begin{align}
  \Nn(P, R) &= \left\{ (I,J)\in \Ee | J \in \hat{\Dd}(P) \backslash \hat{\Dd}(R), I \not\in \hat{\Dd}(P)\right\},\nonumber \\
  \Hh(P, R) &= \left\{ (I,J) \in \Ee | J \in \hat{\Dd}(R), I \in \hat{\Dd}(P) \backslash \hat{\Dd}(R) \right\}.
\end{align}
Similar to mean field and loopy BP, the messages are propagated and updated with the rule in \eqref{apdix:gbp-update-rule} until convergence. Then,
the belief for each region $R$ is given by
\begin{align}
  b_R(\bm{x}_R) \propto \prod_{a\in A_R} \psi_a(\bm{x}_a) \prod_{P\in
  \Pp(R)} m_{P\rightarrow R}(\bm{x}_R) \prod_{D\in \Dd(R)} \prod_{P^{\prime} \in \Pp(D) \backslash \hat{\Dd}(R)} m_{P^{\prime}\rightarrow D}(\bm{x}_D).
\end{align}

\section{Constructing the Root Regions from General Graphs}
\label{apdix:sec:root-region-construct}

\begin{algorithm}[t!]
  \caption{Construct Root Regions from General Graphs.}
  \label{apdix:alg:root-region-general-graph}
  \begin{algorithmic}
    \STATE {\bfseries Input:} Pairwise Markov random field $p(\bm{x})$
    \STATE Draw the factor graph $\Gg_F$ of $p(\bm{x})$
    \STATE Obtain graph $\Gg$ by preserving the variable nodes as they are and converting the factor nodes of $\Gg_F$ into edges
    \STATE Find the subgraph $\Gg_s$ of $\Gg$, such that $\Gg_s$ is planar or complete graph
    \STATE Add the tree-robust basis $\Cc\Bb(\Gg_s)$ of $\Gg_s$ into $\Rr_0$
    \STATE Marked all nodes as \textit{visited} and edged as \textit{used} in $\Gg_s$
    \REPEAT
    \STATE Choose an \textit{unused} edge $e = (s,t)$ from a \textit{visited} node $s$
    \IF{$t$ is visited}
    \STATE Set $\mathrm{path}_1 = e$
    \STATE Find the shortest path $\mathrm{path}_2$ from $s$ to $t$
    via \textit{used} edges
    \ELSE
    \STATE Find path from $s$ to a \textit{visited} $u$ that
    contains edge $e$, this path is set as $\mathrm{path}_1$.
    \STATE Find the shortest path $\mathrm{path}_2$ from $s$ to $u$
    via \textit{used} edges
    \ENDIF
    \STATE Add cycle $C$ consisting of $\mathrm{path}_1$ and
    $\mathrm{path}_2$ to $\Rr_0$.
    \STATE Mark all nodes as \textit{visited} and edges as
    \textit{used} in $C$
    \UNTIL{$\nexists$ \textit{unused} edge $e = (s,t)$ from a
      \textit{visited} node $s$}
  \end{algorithmic}
\end{algorithm}

To explain the concept of \textit{tree-robust} in
\cite{gelfand2012generalized}, we need to explain the concepts of \textit{cycle basis} and \textit{tree exact}, based on which the tree-robust is defined.
\begin{defn}\label{apdix:def:cycle-basis}
  A \textit{cycle basis} of the cycle space of a graph $\Gg$ is a
  set of simple cycles $\Cc\Bb =\left\{ C_1, C_2, \cdots, C_{\mu}
  \right\}$ such that for every cycle $C$ in graph $\Gg$, there
  exists a unique subset $\Cc\Bb_{C} \subseteq \Cc\Bb$ such that the
  set of edges appearing an odd number of times in $\Cc\Bb_C$ comprise the cycle $C$.
\end{defn}

\begin{defn}\label{apdix:def:tree-exact}
  Let $T$ be a spanning tree of graph $\Gg$. A cycle basis $\Cc\Bb$ is
  \textit{tree exact} w.r.t. $T$ if there exists an ordering $\pi$ of
  the cycles in $\Cc\Bb$ such that $\left\{ C_{\pi(i)} \backslash C_{\pi(1)} \cup C_{\pi(2)} \cup \cdots \cup C_{\pi(i-1)} \right\} \neq \emptyset$ for $i=2,\cdots, \mu$.
\end{defn}
Definition~\ref{apdix:def:tree-exact} tells us that if a cycle basis is tree exact w.r.t. $T$ and ordered properly, there is at least one edge of $C_{\pi}$ that has not appeared in any cycles preceding it, and meanwhile, this edge does not appear in the spanning tree $T$.

With the above concepts, we are ready to give the definition of tree-robust.

\begin{defn}\label{apdix:def:tree-robust}
  A cycle basis $\Cc\Bb$ is tree-robust if it is tree exact w.r.t. all spanning trees of $\Gg$.
\end{defn}

Root regions of region graph $\Gg_R$ from planar and compete graphs are explained in Section~\ref{sec:criteria-root-regions}. For general graphs, it basically is to find a subgraph that is a planar or complete graph, and then extract the corresponding tree-robust basis, after which extra cycles are added in by following Algorithm~\ref{apdix:alg:root-region-general-graph}.

\section{Learning of MRFs with Hidden Variables by RENN}
\label{apdix:sec:crf-learning}
For cases where there is a hidden variable $\bm{z}$ apart from the
observable variable $\bm{x}$, denote the joint probability mass
function as
\begin{equation}\label{apdx:eq:joint-px}
  p(\bm{x}, \bm{z}; \bm{\theta}) = \frac{1}{Z(\bm{\theta})} \prod_{a}
  \psi_a(\bm{x}_a, \bm{z}_a; \bm{\theta}_a),
\end{equation}
where $Z(\bm{\theta}) = \sum_{\bm{x}, \bm{z}}\prod_{a} \psi_a(\bm{x}_a,\bm{z}_a;\bm{\theta}_a)$. Since we only have observations for $\bm{x}$, we can only maximize marginalization $p(\bm{x}; \bm{\theta})$ instead the complete joint probability $p(\bm{x}, \bm{z}; \bm{\theta})$. The marginal log-likelihood can be written as
\begin{equation}
  \log{p(\bm{x}; \bm{\theta})}  = \log{Z(\bm{x}; \bm{\theta})} - \log{Z(\bm{\theta})},
\end{equation}
where $Z(\bm{x}; \bm{\theta}) = \sum_{\bm{z}}\prod_{a} \psi_a(\bm{x}_a, \bm{z}_a; \bm{\theta}_a)$.

As discussed in Section~\ref{sec:model-learning-with-renn}, RENN can be used to approximate partition function $F_R(\Bb^{\ast};
\bm{\theta}) \approx - \log{Z(\bm{\theta})}$. We can similarly use a
separate RENN to do approximation $F_R(\Bb^{\ast}_{\bm{x}};
\bm{\theta}) \approx - \log{Z(\bm{x},\bm{\theta})}$, with
$\Bb^{\ast}_{\bm{x}}= \{b_R(\bm{z}_R|\bm{x}; \bm{\omega}^{\ast}), R\in
\Rr\}$. In this case, the corresponding region graph is constructed
from a factor graph with $\bm{x}$ clamped to a given observation.

\newpage

\section{More Experimental Results}
\label{apdx:sec:more-experiments}
In this section, we include more experiment results of inference and learning by RENN in comparison with benchmark methods.

\begin{table*}[t]
  \caption{Inference on grid graph ($\gamma=0.1$). $\ell_1$ error and correlation $\rho$ between true and approximate marginals, and $\log{Z}$ error.}
  \label{apdx:table:infer-grid-gamma0.1}
  \begin{center}
    \begin{small}
      \begin{adjustbox}{width=1\textwidth}
        \begin{tabular}{lcccccccc}
          \toprule
          Metric & $n$ & Mean Field & Loopy BP & Damped BP & GBP & Inference Net & RENN \\
          \midrule
          \multirow{4}{*}{\begin{tabular}[x]{@{}c@{}}$\ell_1$\\error \end{tabular} }
                 &    25   &$0.271 \pm 0.051$ &  $0.086 \pm 0.078$ & $0.084 \pm 0.076$ & $0.057 \pm 0.024$ & $0.111 \pm 0.072$ & \textbf{0.049} $\pm$ 0.078 \\

                 &    100   & $0.283 \pm 0.024$ &  $0.085 \pm 0.041$ & $0.062 \pm 0.024$ & $0.064 \pm 0.019$ & $0.074 \pm 0.034$ & \textbf{0.025} $\pm$ 0.011 \\

                 &    225   & $0.284 \pm 0.019$ &  $0.100 \pm 0.025$ & $0.076 \pm 0.025$ & $0.073 \pm 0.013$ & $ 0.073 \pm 0.012$ & \textbf{0.046} $\pm$ 0.011 \\

                 &    400   & $0.279 \pm 0.014$ &  $0.110 \pm 0.016$ & $0.090 \pm 0.016$ & $0.079 \pm 0.009$ & $ 0.083 \pm 0.009$ & \textbf{0.061} $\pm$ 0.009 \\

          \midrule
          \multirow{4}{*}{\begin{tabular}[x]{@{}c@{}}Corre-\\lation\\ $\rho$ \end{tabular}}
                 &   25    & 0.633 $\pm$ 0.197  &  0.903 $\pm$ 0.114  &  0.905 $\pm$ 0.113  &  0.923 $\pm$ 0.045  &  0.866$\pm$ 0.117 &  \textbf{0.951} $\pm$ 0.112 \\

                 &   100   & 0.582 $\pm$ 0.112  &  0.827 $\pm$ 0.134  &  0.902 $\pm$ 0.059  &  0.899 $\pm$ 0.043  &  0.903$\pm$ 0.049 &   \textbf{0.983} $\pm$ 0.012 \\

                 &   225   & 0.580 $\pm$ 0.080  &  0.801 $\pm$ 0.078  &  0.863 $\pm$ 0.088  &  0.869 $\pm$ 0.037  & 0.873 $\pm$ 0.037 &  \textbf{0.949} $\pm$ 0.022 \\

                 &   400   & 0.596 $\pm$ 0.054  &  0.779 $\pm$ 0.059  &  0.822 $\pm$ 0.047  &  0.852 $\pm$ 0.024  & 0.841 $\pm$ 0.028 &  \textbf{0.912} $\pm$ 0.025 \\

          \midrule
          \multirow{4}{*}{\begin{tabular}[x]{@{}c@{}}$\log{Z}$ \\error\end{tabular}}
                 &   25    & 2.512 $\pm$ 1.060  &  0.549 $\pm$ 0.373  &  0.557 $\pm$ 0.369  &  \textbf{0.169} $\pm$ 0.142  &  0.762 $\pm$ 0.439  &  0.240 $\pm$ 0.140 \\

                 &  100    & 13.09 $\pm$ 2.156  &  1.650 $\pm$ 1.414  &  1.457 $\pm$  1.365 &  \textbf{0.524} $\pm$ 0.313  &  2.836 $\pm$ 2.158  & 1.899 $\pm$ 0.495 \\

                 &  225    & 29.93 $\pm$ 4.679  &  3.348 $\pm$ 1.954  &  3.423 $\pm$ 2.157  &  \textbf{1.008} $\pm$ 0.653  &  3.249 $\pm$ 2.058  & 4.344 $\pm$ 0.813  \\

                 &  400    & 51.81 $\pm$ 4.706  &  5.738 $\pm$ 2.107  &  5.873$\pm$ 2.211   &  \textbf{1.750} $\pm$ 0.869  &  3.953 $\pm$ 2.558  & 7.598 $\pm$ 1.146 \\

                 % \midrule
                 % \multirow{4}{*}{Time}
                 % &   25   &  0.259 $\pm$ 0.076  &  2.990 $\pm$ 4.563  &  1.591 $\pm$ 0.609  &  6.817 $\pm$ 0.339  &  5.253 $\pm$ 1.189  &  5.828 $\pm$ 2.372  \\
                 % &  100   &  1.290 $\pm$ 0.205  &  57.10 $\pm$ 25.64 &  28.70 $\pm$ 24.62 & 46.89$ \pm$ 1.365 & 22.05 $\pm$ 10.53  & 22.63 $\pm$ 6.202 \\

                 % &  225   &  3.838 $\pm$ 1.015  &  184.1 $\pm$ 3.648  &  124.2 $\pm$ 68.70  &  131.1 $\pm$ 3.440  &  36.43 $\pm$ 2.087  & 51.82 $\pm$ 2.431 \\

                 % &  400   &  7.908 $\pm$ 1.508  & 335.3 $\pm$ 12.83   &  326.9 $\pm$ 27.56  &  253.7 $\pm$ 11.00   &  69.54 $\pm$ 5.140  & 103.9 $\pm$ 8.379 \\

          \bottomrule
        \end{tabular}
      \end{adjustbox}
    \end{small}
  \end{center}
  \vskip -0.2in
\end{table*}

\begin{table*}[t]
  \caption{Inference on grid Graph. ($\gamma=1$)}
  \label{apdx:table:infer-grid-gamma1.0}
  \vskip -0.1in
  \begin{center}
    \begin{small}
      \begin{adjustbox}{width=1\textwidth}
        \begin{tabular}{lcccccccc}
          \toprule
          Metric & $n$ & Mean Field & Loopy BP & Damped BP & GBP & Inference Net & RENN \\
          \midrule
          \multirow{4}{*}{\begin{tabular}[x]{@{}c@{}}$\ell_1$\\error \end{tabular} }
                 & 25   &  0.131 $\pm$ 0.080  &  \textbf{0.022} $\pm$ 0.017  &  0.022 $\pm$ 0.018  &  0.137 $\pm$ 0.026  &  0.043 $\pm$ 0.017  &  0.027 $\pm$ 0.014 \\
                 & 100  &  0.130 $\pm$ 0.041  &  0.025 $\pm$ 0.014  &  0.025 $\pm$ 0.014  &  0.146 $\pm$ 0.020  &  0.046 $\pm$ 0.009  &  \textbf{0.017} $\pm$ 0.002  \\

                 &225   &  0.135 $\pm$ 0.024  &  0.024 $\pm$ 0.010  &  0.023 $\pm$ 0.009  &  0.154 $\pm$ 0.012  &  0.052 $\pm$ 0.010  &  \textbf{0.017} $\pm$ 0.003 \\

                 &400   &  0.131 $\pm$ 0.020  &  0.020 $\pm$ 0.003  &  0.020 $\pm$ 0.003  &  0.158 $\pm$ 0.007  &  0.052 $\pm$ 0.007  &  \textbf{0.017} $\pm$ 0.001  \\

          \midrule
          \multirow{4}{*}{\begin{tabular}[x]{@{}c@{}}Corre-\\lation \\$\rho$\end{tabular}}
                 & 25   &  0.849 $\pm$ 0.159  &  \textbf{0.992} $\pm$ 0.011  &  0.991 $\pm$ 0.012  &  0.798 $\pm$ 0.088  &  0.980 $\pm$ 0.015  & 0.988 $\pm$ 0.025  \\
                 & 100  &  0.841 $\pm$ 0.087  &  0.988 $\pm$ 0.013  &  0.988 $\pm$ 0.012  &  0.788 $\pm$ 0.051  &  0.976 $\pm$ 0.013  &  \textbf{0.997} $\pm$0.001 \\

                 & 225  &  0.824 $\pm$ 0.057  &  0.989 $\pm$ 0.010  &  0.990 $\pm$ 0.010  &  0.764 $\pm$ 0.022  &  0.966 $\pm$ 0.016  &  \textbf{0.996} $\pm$ 0.001 \\

                 & 400  &  0.828 $\pm$ 0.043  &  0.993 $\pm$ 0.002  &  0.993 $\pm$ 0.002  &  0.759 $\pm$ 0.018  &  0.967 $\pm$ 0.013  &  \textbf{0.997} $\pm$ 0.001  \\

          \midrule
          \multirow{4}{*}{\begin{tabular}[x]{@{}c@{}}$\log{Z}$ \\error\end{tabular}}
                 & 25  &  2.113 $\pm$ 1.367  &  \textbf{0.170} $\pm$ 0.199  &  0.194 $\pm$ 0.188  &  0.605 $\pm$ 0.611  &  2.214 $\pm$ 0.775  &  0.649 $\pm$ 0.363  \\

                 &100  &  8.034 $\pm$ 2.523  &  \textbf{0.372} $\pm$ 0.427  &  0.415 $\pm$ 0.422  &  1.545 $\pm$ 1.081  &  11.14 $\pm$ 0.954  &  3.129 $\pm$ 0.520  \\

                 &225  &  17.923 $\pm$ 3.474 &  0.952 $\pm$ 1.037  &  \textbf{0.917} $\pm$ 0.922  &  3.143 $\pm$ 2.122  &  25.55 $\pm$ 2.025  &  7.473 $\pm$ 0.906  \\

                 &400  &  31.74 $\pm$ 4.766          &  \textbf{0.919} $\pm$ 0.684   &  1.011 $\pm$ 0.685  &  3.313 $\pm$ 1.872  &  46.61 $\pm$ 3.094  &  12.77 $\pm$ 0.991  \\

                 % \midrule
                 % \multirow{4}{*}{Time}
                 % & 25  &  0.281 $\pm$ 0.129  &  0.937 $\pm$ 0.442  &  1.253 $\pm$ 0.403  &  6.982 $\pm$ 0.260  &  5.146 $\pm$ 0.398  &  5.341 $\pm$ 1.368  \\

                 % & 100 &  1.219 $\pm$ 0.314  &  6.339 $\pm$ 3.965  &  6.948 $\pm$ 1.664  &  49.64 $\pm$ 1.301  &  21.59 $\pm$ 4.703  &  31.79 $\pm$ 15.51  \\

                 % &125  &  3.539 $\pm$ 1.107  &  33.09 $\pm$ 48.62  &  17.85 $\pm$ 4.765  &  127.1 $\pm$ 3.547  &  56.99 $\pm$ 4.625  &  94.05 $\pm$ 38.64  \\

                 % &400  &  6.313 $\pm$ 1.465  &  49.26 $\pm$ 65.20  &  33.07 $\pm$ 8.777  &
                 % 255.1 $\pm$ 10.13  &  157.3 $\pm$ 34.04  &  172.58 $\pm$ 27.08  \\
          \bottomrule
        \end{tabular}
      \end{adjustbox}
    \end{small}
  \end{center}
  \vskip -0.2in
\end{table*}

\subsection{More Inference Results on Grid Graphs}
\label{apdx:sec:extra-grid-results}
This section includes additional experimental comparisons for inference on grid graphs as supplementary for Section~\ref{sec:inference-grid}.
Experiments are carried out with the standard deviation of $\{h_i\}$ in setting $\gamma=0.1$ and $\gamma=1$, which reflects the relative strength of univariate log-potentials to pairwise log-potentials. More grid sizes are also shown here.

\begin{table}[h]
  \caption{Inference with the \textit{infinite face} on grid, $n=25$.}
  \label{apdx:tab:infer-infinite-face}
  \begin{center}
    \begin{small}
      \begin{sc}
        \begin{tabular}{llcc}
          \toprule
          $\gamma$ & Metric & GBP & RENN \\
          \midrule
          \multirow{3}{*}{0.1}
                   & $\ell_1$ Error & 0.061 $\pm$ 0.025 & \textbf{0.025} $\pm$ 0.020 \\

                   & $\rho$   & 0.913 $\pm$ 0.049  &  \textbf{0.984} $\pm$ 0.021  \\
                   & $\log{Z}$ Error & 3.564 $\pm$ 2.823  &  0.384 $\pm$ 0.223  \\
          \midrule
          \multirow{3}{*}{1}
                   & $\ell_1$ Error & 0.145 $\pm$ 0.028  & \textbf{0.016} $\pm$ 0.010 \\

                   & $\rho$   & 0.783 $\pm$ 0.091  &  \textbf{0.995} $\pm$ 0.010 \\
                   & $\log{Z}$ Error & 0.825 $\pm$ 0.841  & 0.364 $\pm$ 0.201 \\

          \bottomrule
        \end{tabular}
      \end{sc}
    \end{small}
  \end{center}
  \vskip -0.2in
\end{table}

\begin{table*}[hbt!]
  \caption{Inference on complete graph of size $9$.}
  \label{apdx:tab:infer-full-n9}
  \vskip -0.1in
  \begin{center}
    \begin{small}
      \begin{adjustbox}{width=1\textwidth}
        \begin{tabular}{lcccccccc}
          \toprule
          Metric & $\gamma$ & Mean Field & Loopy BP & Damped BP & GBP & Inference Net & RENN \\
          \midrule
          \multirow{4}{*}{\begin{tabular}[x]{@{}c@{}}$\ell_1$\\error\end{tabular}}
          % pen1
                 & 0.1   &  0.294 $\pm$ 0.061  &  0.120 $\pm$ 0.038  &  0.118 $\pm$ 0.034  &   0.237 $\pm$ 0.061  &  \textbf{0.109} $\pm$ 0.025  &  0.130 $\pm$ 0.085  \\

                 % pen2
                 & 1     &  0.233 $\pm$ 0.133  &  0.200 $\pm$ 0.098  &  0.201 $\pm$ 0.098  &  0.246 $\pm$ 0.135  &  0.196 $\pm$ 0.061  &  \textbf{0.137} $\pm$ 0.117  \\

                 % pen3
                 & 2     &  0.187 $\pm$ 0.131  &  0.176 $\pm$ 0.114  &  0.177 $\pm$ 0.113  &  0.247 $\pm$ 0.117  &  0.182 $\pm$ 0.084  &  \textbf{0.067} $\pm$ 0.045  \\

                 % pen3
                 & 3     &  0.155 $\pm$ 0.120  &  0.145 $\pm$ 0.112  &  0.146 $\pm$ 0.112  &  0.204 $\pm$ 0.107  &  0.152 $\pm$ 0.079  &  \textbf{0.060} $\pm$ 0.038  \\

                 % pen3
                 & 4     &  0.124 $\pm$ 0.115  &  0.120 $\pm$ 0.103  &  0.121 $\pm$ 0.102  &  0.194 $\pm$ 0.076  &  0.129 $\pm$ 0.071  &  \textbf{0.051} $\pm$ 0.050  \\

          \midrule
          \multirow{4}{*}{\begin{tabular}[x]{@{}c@{}}Corre-\\lation\\$\rho$\end{tabular}}
                 & 0.1   & 0.262 $\pm$ 0.177  &  0.695 $\pm$ 0.104  &  0.698 $\pm$ 0.099  &  0.446 $\pm$ 0.196  &   0.720 $\pm$ 0.065  &  \textbf{0.741} $\pm$ 0.220  \\

                 & 1     & 0.465 $\pm$ 0.349  &  0.538 $\pm$ 0.292  &  0.538 $\pm$ 0.292  &  0.461 $\pm$ 0.331  &  0.639 $\pm$ 0.159   &  \textbf{0.769} $\pm$ 0.313  \\

                 &  2    & 0.587 $\pm$ 0.300  &  0.619 $\pm$ 0.284  &  0.619 $\pm$ 0.282  &  0.457 $\pm$ 0.257  &   0.645 $\pm$ 0.175  &  \textbf{0.929} $\pm$ 0.118  \\

                 &  3    & 0.657 $\pm$ 0.289  &  0.697 $\pm$ 0.267  &  0.697 $\pm$ 0.265  &  0.582 $\pm$ 0.218  &  0.697 $\pm$ 0.162   &  \textbf{0.936} $\pm$ 0.076  \\

                 &  4    & 0.758 $\pm$ 0.257  &   0.778 $\pm$ 0.221 &  0.776 $\pm$ 0.221  &  0.597 $\pm$ 0.177  &  0.753 $\pm$ 0.178   &  \textbf{0.941} $\pm$ 0.099  \\

          \midrule
          \multirow{4}{*}{\begin{tabular}[x]{@{}c@{}}$\log{Z}$ \\error\end{tabular}}
                 & 0.1  & 8.402 $\pm$ 4.369  &  34.61 $\pm$ 2.439  &  34.74 $\pm$ 2.195  &  \textbf{1.763} $\pm$ 1.176  &  35.46 $\pm$ 1.651  &  3.171 $\pm$ 1.259   \\

                 &  1   & 6.473 $\pm$ 3.737  &  45.91 $\pm$ 6.888  &  45.96 $\pm$ 6.927  &  \textbf{1.826} $\pm$ 2.024  &  51.87 $\pm$ 6.150  &  2.796 $\pm$ 1.194  \\

                 &  2   & 5.830 $\pm$ 2.979  &  75.35 $\pm$ 14.58  &  75.46 $\pm$ 14.57  &  3.080  $\pm$ 2.958  &  81.23 $\pm$ 12.939 &  \textbf{2.577} $\pm$ 1.845  \\

                 &  3   & 4.401 $\pm$ 2.522  &  111.0 $\pm$ 22.20  &  111.1 $\pm$ 22.17  &  3.205 $\pm$ 3.720  &  116.1 $\pm$ 19.76  &  \textbf{2.645} $\pm$ 1.507  \\

                 &  4   & 3.037 $\pm$ 2.122  &  142.9 $\pm$ 25.58  &  143.1 $\pm$ 25.56  &  5.167 $\pm$ 5.249  &  147.2 $\pm$ 23.38  &  \textbf{1.820} $\pm$ 1.306  \\

                 % \midrule
                 % \multirow{4}{*}{Time}
                 % & 0.1  &  0.270 $\pm$ 0.096  &  0.490 $\pm$ 0.315  &  0.590 $\pm$ 0.544  & 3.023 $\pm$ 0.122  &  4.644 $\pm$ 0.326  &  4.979 $\pm$ 1.612  \\

                 % &  1   & 0.186 $\pm$ 0.059   &  0.302 $\pm$ 0.072  &  0.501 $\pm$ 0.112  &  2.854 $\pm$ 0.098  &  4.330 $\pm$ 0.423  &  4.905 $\pm$ 1.815  \\

                 % &  2   & 0.111 $\pm$ 0.031   &  0.178 $\pm$ 0.028  &  0.315 $\pm$ 0.048  &  2.722 $\pm$ 0.099  &  4.529 $\pm$ 0.569  &  5.329 $\pm$ 1.725  \\

                 % &  3   & 0.112 $\pm$ 0.072   &  0.155 $\pm$ 0.025  &  0.281 $\pm$ 0.057  &  2.671 $\pm$ 0.131  &  4.667 $\pm$ 1.217  &  4.477 $\pm$ 2.076  \\

                 % &  4   & 0.098 $\pm$ 0.039   &  0.141 $\pm$ 0.026  &  0.279 $\pm$ 0.062  &  2.731 $\pm$ 0.110  &  5.091 $\pm$ 1.798  &  4.309 $\pm$ 1.152  \\

          \bottomrule
        \end{tabular}
      \end{adjustbox}
    \end{small}
  \end{center}
\end{table*}

\begin{table*}[t]
  \caption{Inference on complete graph of size $16$.}
  \label{apdx:tab:infer-full-n16}
  % \vskip -0.05in
  \begin{center}
    \begin{small}
      \begin{adjustbox}{width=1\textwidth}
        \begin{tabular}{lcccccccc}
          \toprule
          Metric & $\gamma$ & Mean Field & Loopy BP & Damped BP & GBP & Inference Net & RENN \\
          \midrule
          \multirow{4}{*}{\begin{tabular}[x]{@{}c@{}}$\ell_1$-\\error\end{tabular}}
          % pen3
                 & 0.1   &  0.303 $\pm$ 0.056  &  0.176 $\pm$ 0.039  &  0.174 $\pm$ 0.038  &  0.244 $\pm$ 0.047  &  0.174 $\pm$ 0.044  &  \textbf{0.169} $\pm$ 0.052  \\

                 % pen2
                 &  1    &  0.273 $\pm$ 0.086  &  0.239 $\pm$ 0.059  &  0.239 $\pm$ 0.059  &  0.260 $\pm$ 0.086  &  0.249 $\pm$ 0.067  &  \textbf{0.181} $\pm$ 0.092  \\

                 % pen2
                 &  2    &  0.231 $\pm$ 0.079  &  0.222 $\pm$ 0.064  &  0.221 $\pm$ 0.064  &  0.249 $\pm$ 0.078  &  0.232 $\pm$ 0.069  &  \textbf{0.170} $\pm$ 0.109  \\

                 % pen3
                 &  3    &  0.218 $\pm$ 0.042  &  0.204 $\pm$ 0.038  &  0.204 $\pm$ 0.038  &  0.247 $\pm$ 0.065  &  0.213 $\pm$ 0.051  &  \textbf{0.138} $\pm$ 0.106  \\

                 % pen1
                 &  4    &  0.197 $\pm$0.049   &  0.181 $\pm$ 0.035  &  0.180 $\pm$ 0.034  &  0.210 $\pm$ 0.070  &  0.174 $\pm$ 0.030  &  \textbf{0.125} $\pm$ 0.050  \\

          \midrule
          \multirow{4}{*}{\begin{tabular}[x]{@{}c@{}}Corre-\\lation\\$\rho$\end{tabular}}
                 & 0.1   &  0.231 $\pm$ 0.196  &  0.509 $\pm$ 0.056  &  0.510 $\pm$ 0.055  &  0.316 $\pm$ 0.207  &  0.506 $\pm$ 0.063  &  \textbf{0.539} $\pm$ 0.235  \\

                 &  1    &  0.381 $\pm$ 0.255  &  0.514 $\pm$ 0.185  &  0.515 $\pm$ 0.185  &  0.445 $\pm$ 0.223  &  0.533 $\pm$ 0.150  &  \textbf{0.756} $\pm$ 0.187  \\

                 &  2    &  0.535 $\pm$ 0.207  &  0.569 $\pm$ 0.180  &  0.570 $\pm$ 0.179  &  0.480 $\pm$ 0.186  &  0.559 $\pm$ 0.176  &  \textbf{0.750} $\pm$ 0.261  \\

                 &  3    &  0.586 $\pm$ 0.142  &  0.618 $\pm$ 0.134  &  0.619 $\pm$ 0.134  &  0.502 $\pm$ 0.144  &  0.613 $\pm$ 0.128  &  \textbf{0.853} $\pm$ 0.159  \\

                 &  4    &  0.622 $\pm$ 0.166  &  0.658 $\pm$ 0.133  &  0.660 $\pm$ 0.132  &  0.564 $\pm$ 0.165  &  0.693 $\pm$ 0.060  &  \textbf{0.868} $\pm$ 0.053  \\

          \midrule
          \multirow{4}{*}{\begin{tabular}[x]{@{}c@{}}$\log{Z}$ \\error\end{tabular}}
                 & 0.1   &  24.45 $\pm$ 7.560  &  143.7 $\pm$ 9.297  &  145.5 $\pm$ 6.096  &  166.3 $\pm$ 11.98  &  148.5 $\pm$ 3.522  &  \textbf{12.57} $\pm$ 3.689  \\

                 &  1    &  20.66 $\pm$ 5.451  &  178.7 $\pm$ 22.18  &  178.9 $\pm$ 21.88  &  153.3 $\pm$ 25.29  &  213.6 $\pm$ 12.75  &  \textbf{14.41} $\pm$ 4.135  \\

                 &  2    &  16.04 $\pm$ 4.352  &  296.3 $\pm$ 44.41  &  296.9 $\pm$ 44.24  &  116.9 $\pm$ 32.72  &  335.1 $\pm$ 32.86  &  \textbf{13.37} $\pm$ 4.531  \\

                 &  3    &  13.87 $\pm$ 6.554  &  432.7 $\pm$ 66.44  &  433.4 $\pm$ 66.30  &  100.2 $\pm$ 39.62  &  462.9 $\pm$ 53.61  &  \textbf{12.56} $\pm$ 6.046  \\

                 &  4    &  \textbf{10.74} $\pm$ 7.385  &  565.7 $\pm$ 73.33  &  566.1 $\pm$ 73.13  &  106.0 $\pm$ 54.43  &  588.3 $\pm$ 62.58  &  14.72 $\pm$ 4.155  \\

                 % \midrule
                 % \multirow{4}{*}{Time}
                 % & 0.1   &  0.555 $\pm$ 0.135  &  1.705 $\pm$ 1.256  &  1.428 $\pm$ 0.400  &  13.68 $\pm$ 0.432  &  12.63 $\pm$ 0.592  &  15.26 $\pm$ 4.759  \\

                 % &  1    &  0.457 $\pm$ 0.103  &  0.777 $\pm$ 0.167  &  1.285 $\pm$ 0.190  &  14.29 $\pm$ 0.487  &  12.45 $\pm$ 0.603  &  16.16 $\pm$ 5.695  \\

                 % &  2    &  0.424 $\pm$ 0.101  &  0.630 $\pm$ 0.096  &  1.143 $\pm$ 0.196  &  14.76 $\pm$ 0.512  &  15.04 $\pm$ 1.106  &  14.74 $\pm$ 4.606  \\

                 % &  3    &  0.377 $\pm$ 0.114  &  0.567 $\pm$ 0.118  &  1.084 $\pm$ 0.254  &  14.49 $\pm$ 0.392  &  15.62 $\pm$ 1.218  &  13.40 $\pm$ 3.070  \\

                 % &  4    &  0.521 $\pm$ 0.475  &  0.543 $\pm$ 0.137  &  1.018 $\pm$ 0.171  &  14.00 $\pm$ 0.379  &  14.59 $\pm$ 5.254  &  14.24 $\pm$ 2.536 \\

          \bottomrule
        \end{tabular}
      \end{adjustbox}
    \end{small}
  \end{center}

\end{table*}

The results reported in Table~\ref{apdx:table:infer-grid-gamma0.1} and \ref{apdx:table:infer-grid-gamma1.0} here give a richer comparison for inference on grid graphs. In all cases except one, beliefs of RENN outperform benchmark algorithms with large marginals. As expected, performances of loopy BP and its variant damped BP are similar in general while damped BP sometimes gets better estimations. Both loopy BP and damped BP have better marginal estimations than the mean field method in all of our considered scenarios. GBP and Inference Net outperform loopy BP and damped BP at case $\gamma=0.1$, but fall behind in case of $\gamma=1$ in general. RENN shows superior performance in most cases.

With regard to the error of partition function values, GBP gets the most accurate estimations when $\gamma=0.1$. $\log{Z}$ estimated by loopy BP and damped BP is better for $\gamma=1$. Partition function estimation by RENN is competitive for different considered cases.

The region graphs in this set of experiments uses all faces of grid graphs but the \textit{infinite face} (the perimeter circle). {For instance, the region $\{1, 2, 3, 4, 5, 6, A, B, C ,E, F, G\}$ is obtained from the infinite face in the 2-by-3 grid in Figure~\ref{fig:factor-region-graphs} in the paper}. By comparing Table~\ref{apdx:tab:infer-infinite-face} with the $n=25$ cases of Table~\ref{apdx:table:infer-grid-gamma0.1} and \ref{apdx:table:infer-grid-gamma1.0}, performance of RENN can be further improved when we include the infinite face in building region graphs from grids. On the contrary, performance of GBP drops slightly after including the infinite face. But number of nodes in the region built from the infinite face would scale with the perimeter of grid graph. Since RENN already has reasonably good accuracy outperforming benchmark methods as shown in Table~\ref{apdx:table:infer-grid-gamma0.1} and \ref{apdx:table:infer-grid-gamma1.0}, we suggest to drop the infinite face in constructing region graphs from grids.

            %             \begin{table}[t]
            %             \caption{Consumed time per epoch (unit second).}
            %             \label{tab:time-training}
            %             \begin{center}
            %             \begin{small}
            %             \begin{sc}
            %             \begin{tabular}{lcc}
            %             \toprule
            %             $n$ & 25 & 100 \\
            %             \midrule
            %             Mean Field & 8.850 & 24.36 \\
            %             Loopy BP &  41.58 & 94.97 \\
            %             Damped BP & 35.85 & 156.8 \\
            %             GBP &  1.466 & 9.245  \\
            %             Inference Net & 1.466 & 5.314 \\
            %             RENN &  2.329 & 10.98\\

            %             \bottomrule
            %           \end{tabular}
            %             \end{sc}
            %             \end{small}
            %             \end{center}
            %             \vskip -0.2in
            %             \end{table}

\subsection{More Inference Results on Complete Graphs}
\label{apdx:sec:extra-result-complete}
This section provides additional inference results on complete graphs as supplementary content for Section~\ref{apdx:sec:infer-complete-graph}.
We carry out the inference experiments on complete graphs of size $n=9$ and $n=16$. For each graph size, setting of $\gamma$ includes $\left\{ 0.1, 1, 2, 3, 4 \right\}$.

RENN outperforms all other benchmark methods except for one case at $\gamma=0.1$ in size-$9$ graph, as shown in Table~\ref{apdx:tab:infer-full-n9} and \ref{apdx:tab:infer-full-n16}. In the case of $\gamma=0.1$ in Table~\ref{apdx:tab:infer-full-n9}, Inference Net outperforms RENN slightly w.r.t. $\ell_1$ error, i.e. $0.109$ versus $0.130$, but falls behind RENN w.r.t. correlation $\rho$ ($0.720$ versus $0.741$) and $\log{Z}$ estimation significantly ($35.46$ versus $3.171$).

In the complete graphs, GBP does not have an advantage over loopy BP and damped BP any more, RENN operating on the same region graphs as those for GBP, gives consistently better marginal distribution estimates. Also, generally speaking, the performance of Inference Net is close to loopy BP and damped in most cases of complete graphs.

As for partition function evaluations of complete graphs, the results are quite different from those of grid graphs, by observing Table~\ref{apdx:tab:infer-full-n9} and \ref{apdx:tab:infer-full-n16}. Loopy BP, damped BP, and Inference Net are getting very large errors of partition function as univariate log-potentials are more different from each other, i.e. $\gamma$ gets larger. GBP has reasonable good estimation of $\log{Z}$ in smaller sized complete graph in Table~\ref{apdx:tab:infer-full-n9}, but gets large $\log{Z}$ error in a bit larger complete graph as in Table~\ref{apdx:tab:infer-full-n16}. Mean field methods give a much better estimation of $\log{Z}$ in complete graphs than loopy BP, damped BP, and Inference Net, but it has poorer marginal distribution estimations.

\subsection{Further Discussion on the Hyperparameter $\lambda$}
\label{apdx:sec:discussion-lambda}
\begin{table*}[t]
  \caption{Inference with \textit{ill} setting $\lambda=0$ on Grid Graphs ($\gamma=1$).}
  \label{apdx:tab:lambda0-inference-grid}
  \centering
  \begin{small}
    \begin{tabular}{llcc}
      \toprule
      $n$ & Metric & Inference Net & RENN \\
      \midrule
      \multirow{3}{*}{25}
          & $\ell_1$ Error & 0.091 $\pm$ 0.026 & 0.107 $\pm$ 0.032\\
          & Correlation $\rho$ & 0.908 $\pm$ 0.050 & 0.879 $\pm$ 0.074  \\
          & $\log{Z}$ Error & 3.304 $\pm$ 1.111 & 2.256 $\pm$ 1.006 \\
      \midrule
      \multirow{3}{*}{100}
          & $\ell_1$ & 0.098 $\pm$ 0.021 & 0.108 $\pm$ 0.021 \\
          & $\rho$   & 0.893 $\pm$ 0.049 & 0.874 $\pm$ 0.044  \\
          & $\log{Z}$ Error & 16.68 $\pm$ 1.245 & 12.29 $\pm$ 1.748 \\

      \bottomrule
    \end{tabular}
  \end{small}
\end{table*}

\begin{table*}[t]
  \caption{Inference with \textit{ill} setting $\lambda=0$ on Complete Graphs ($\gamma=1$).}
  \label{apdx:tab:lambda0-inference-complete}
  \centering
  \begin{small}
    \begin{tabular}{llcc}
      \toprule
      $n$ & Metric & Inference Net & RENN \\
      \midrule
      \multirow{3}{*}{9}
          & $\ell_1$ Error & 0.198 $\pm$ 0.058 & 0.155 $\pm$ 0.083\\
          & Correlation $\rho$ & 0.614 $\pm$ 0.157 & 0.691 $\pm$ 0.245 \\
          & $\log{Z}$ Error & 52.49 $\pm$ 6.181 & 4.228 $\pm$ 1.280 \\
      \midrule
      \multirow{3}{*}{16}
          & $\ell_1$ & 0.253 $\pm$ 0.063 & 0.192 $\pm$ 0.082 \\
          & $\rho$   & 0.501 $\pm$ 0.140 & 0.686 $\pm$ 0.176  \\
          & $\log{Z}$ Error & 215.7 $\pm$ 12.75 & 20.07 $\pm$ 4.115 \\
      \bottomrule
    \end{tabular}
  \end{small}
\end{table*}

Our experiments chose hyperparameter $\lambda$ heuristically from $\left\{1 ,3 ,5, 10\right\}$. Heuristic methods for hyperparameter selection are common practice. Alternatively, it is also feasible to apply the setting of $\lambda$ from controlled experiments to a similar class of problems. For cases where the above two options can not be used, one may calibrate hyperparameter $\lambda$ of a RENN with referring to an upper or lower bound of the log-partition (e.g., the lower bound of $\log{Z}$ provided by mean field methods).

Although it has been motivated analytically in Section~\ref{sec:infer-renn} in our paper, we show here the necessity of the regularization parameter $\lambda$ by numerical results coming from the ill-setting of $\lambda$. We run the inference by both RENN and benchmark Inference Net on both grid and complete graphs with $\lambda=0$, i.e. without enforcing the regularization on belief consistency. The results are shown in Table~\ref{apdx:tab:lambda0-inference-grid} and \ref{apdx:tab:lambda0-inference-complete}. By comparing the Table~\ref{apdx:tab:lambda0-inference-grid} with corresponding items in Table~\ref{apdx:table:infer-grid-gamma1.0}, it can be seen that both RENN and Inference Net show a degenerated inference performance, i.e., larger $\ell_1$ and $\log{Z}$ errors and smaller correlation in evaluation against the true $p(x_i)$ and $p(x_i,x_j)$. The comparison confirms our analysis on regulating the belief consistency in Section~\ref{sec:infer-renn} in the paper. This ill-setting comparison also shows that belief consistency level is indicated by the selected metrics ($\ell_1$ errors and correlation $\rho$), since removing the enforcement of the regularization degenerates the performance, and
the true univariate and pairwise marginals in assessments are certainly consistent. For the complete graphs, the similar phenomena can be observed by comparing Table~\ref{apdx:tab:lambda0-inference-complete} to corresponding items of Table~\ref{apdx:tab:infer-full-n9} and \ref{apdx:tab:infer-full-n16}.

\subsection{MRF Learning with Inference of RENN}
\label{apdx:sec:extrain-learning-time}

\begin{table}[!t]
  \caption{Average consumed time per epoch (unit: second) for MRF learning cases in Table~\ref{tab:nll-training-grid-complete}.}
  \label{apdx:tab:time-training}
  \begin{center}
    \begin{small}
      \begin{sc}
        \begin{tabular}{lccccc}
          \toprule
          {Graph Type} & \multicolumn{3}{c}{Grid Graph} & \multicolumn{2}{c}{Complete Graph} \\
          \midrule

          {$n$} & 25  &  100  &  225  &  9  &  16  \\
          \midrule
          Mean Field    &  8.850  &  24.36  &  40.09  &  0.838  &  2.449 \\
          Loopy BP      &  41.58  &  94.97  &  335.1  &  1.341  &  12.40  \\
          Damped BP     &  35.85  &  156.8  &  525.1  &  1.649  &  5.431  \\
          GBP           &  1.997  &  9.245  &  12.37  &  0.424  &  1.387  \\
          Inference Net &  1.436  &  2.553  &  19.49  &  0.289  &  0.882  \\
          RENN          &  1.371  &  5.757  &  16.03  &  0.846  &  2.262  \\
          \bottomrule
        \end{tabular}
      \end{sc}
    \end{small}
  \end{center}
  \vskip -0.2in
\end{table}

The training and testing datasets for MRF learning are also attached to our code in the supplementary files. The datasets are sampled by using a variant of forward filtering backward sampling method, the implementation of which is available in our code.

In Table~\ref{apdx:tab:time-training} here, we report the average time consumed per epoch of MRF learning in other cases which corresponding the Table~\ref{tab:nll-training-grid-complete} in our paper. As the size of graphs increases, the advantage of RENN w.r.t. computation time gets more significant. Meanwhile, learning of MRFs by using inference of RENN helps MRF models fit data better as shown in Table~\ref{tab:nll-training-grid-complete} in our paper.

\end{document}